\documentclass{article} 
\usepackage[preprint]{colm2026_conference}
\usepackage{wrapfig}
\usepackage{booktabs}
\usepackage{multirow}
\usepackage{longtable}
\usepackage{graphicx}
\usepackage{siunitx}
\usepackage{subcaption}
\usepackage{amsmath}
\usepackage[dvipsnames,table]{xcolor}
\usepackage{array}
\usepackage{tabularx}
\usepackage{threeparttable}
\usepackage{natbib}
\usepackage[most]{tcolorbox}
\tcbuselibrary{skins}
\usepackage{url}
\usepackage{makecell}
\usepackage{colortbl}
\usepackage{algorithm}
\usepackage{algpseudocode}
\usepackage{microtype}
\usepackage[hidelinks]{hyperref}
\usepackage{lineno}

\usepackage{amsmath,amsfonts,bm}

\def\eqref#1{equation~\ref{#1}}

\def\1{\bm{1}}

\DeclareMathAlphabet{\mathsfit}{\encodingdefault}{\sfdefault}{m}{sl}
\SetMathAlphabet{\mathsfit}{bold}{\encodingdefault}{\sfdefault}{bx}{n}

\definecolor{LightSkyBlue}{RGB}{135,206,250}
\definecolor{LightRed}{RGB}{255, 230, 230}
\definecolor{LightRoyalBlue}{RGB}{236, 240, 252}
\definecolor{LightGreen}{RGB}{230, 242, 230}
\definecolor{LightOrange}{RGB}{255, 246, 230}
\definecolor{LightPurple}{RGB}{242, 230, 242}
\definecolor{LightBrown}{RGB}{246, 234, 234}
\definecolor{LighterSkyBlue}{RGB}{243, 250, 255}
\definecolor{LightLimeGreen}{RGB}{235, 250, 235}
\definecolor{LightPeach}{RGB}{255, 251, 248}
\definecolor{LightPink}{RGB}{255, 249, 250}
\definecolor{LightThistle}{RGB}{251, 249, 251}
\definecolor{darkblue}{rgb}{0, 0, 0.5}
\hypersetup{colorlinks=true, citecolor=darkblue, linkcolor=darkblue, urlcolor=darkblue}

\newcolumntype{L}[1]{>{\raggedright\arraybackslash}p{#1}}
\newcolumntype{Y}{>{\raggedright\arraybackslash}X}

\newtcolorbox{promptbox}[1]{
    enhanced, colback=gray!5!white, colframe=gray!75!black,
    fonttitle=\bfseries, title=#1, coltitle=black,
    attach boxed title to top left={yshift=-2mm, xshift=3mm},
    boxed title style={colback=white, sharp corners}, boxsep=1mm
}
\newtcolorbox{clusterbox}[3]{
    enhanced, colback=#2!10!white, colframe=#2!80!black,
    fonttitle=\bfseries, coltitle=black, title={#1: #3},
    attach boxed title to top left={yshift=-0.25cm, xshift=0.5cm},
    boxed title style={colback=white, colframe=#2!80!black}
}
\tcbset{
  pipelinebox/.style={
    enhanced, sharp corners, boxrule=0.6pt,
    colframe=blue!50!black, colback=white, title filled,
    colbacktitle=blue!4!white, coltitle=black, fonttitle=\bfseries\sffamily,
    attach boxed title to top left={xshift=2mm,yshift*=-2mm},
    boxed title style={sharp corners, boxrule=0pt, interior style={fill=blue!4!white}},
    drop shadow=black!15, left=2mm, right=2mm, top=1.2mm, bottom=1.2mm
  }
}

\title{{AInstein: Can LLMs Solve Research Problems From\\
Parametric Memory Alone?}}

\author{%
\parbox{\linewidth}{\centering
Shambhavi Mishra$^{*1,4,8}$ \&
Gaurav Sahu$^{*2,3,4}$ \\
Marco Pedersoli$^{1,2,8}$ \quad Laurent Charlin$^{2,3,5,6}$ \quad
Jose Dolz$^{1,8}$ \quad Christopher Pal$^{2,4,5,6,7}$
} \\
\parbox{\linewidth}{\centering
$^{1}$LIVIA, ÉTS Montréal \quad $^{2}$Mila -- Quebec AI Institute \quad $^{3}$HEC Montréal \\ $^{4}$ServiceNow Research
 \quad $^{5}$Canada CIFAR AI Chair \\ $^{6}$Université de Montréal \quad $^{7}$Polytechnique Montréal \\
$^{8}$International Laboratory on Learning Systems (ILLS) \\
$^{*}$Equal Contribution}
}

\begin{document}

\ifcolmsubmission
\linenumbers
\fi

\maketitle

\begin{abstract}
Can large language models solve AI research problems using only their parametric knowledge, without fine-tuning, retrieval, or other external aids? We introduce \textbf{AInstein}, a framework for testing whether LLM agents can generate and refine solutions to research problems through iterative critique loops. A blind study with 20 domain experts on held-out ICLR~2026 problems validates our automated metrics, which we then scale to 1,214 ICLR~2025 papers using an LLM-as-a-judge paradigm. Two metrics capture complementary aspects of performance: \textbf{Success Rate} (does the solution address the problem?) and \textbf{Rediscovery} (does it match the published approach?).
LLMs succeed on over 70\% of problems, yet strictly rediscover the published solution less than 19\% of the time, suggesting genuine problem-solving rather than associative recall. However, this ability has clear limits: models handle familiar methodological territory well but fail when solutions require cross-domain analogical transfer, a pattern we call the \textit{parametric knowledge boundary}. On the ResearchPlanGen benchmark (2,645 problems), our training-free iterative refinement strategy matches RL finetuning, and a criteria-coverage analysis pins down the ceiling of what test-time refinement alone can achieve. Together, these findings map both the capabilities and the limits of LLMs as autonomous scientific problem-solvers.
\end{abstract}

\section{Introduction}
Large language models (LLMs) encode vast scientific knowledge in their parameters, but can they \textit{use} it to solve research problems, or only recall fragments of it \citep{Xu2025,chollet2019measureintelligence,zhang2025advancingscientificmethodlarge,min-etal-2022-rethinking}?
LLMs already show strong performance in mathematics, programming, and research workflows \citep{fang2024mathodysseybenchmarkingmathematicalproblemsolving,wang2024scibenchevaluatingcollegelevelscientific,zheng2025knowledgeaugmentedcomplexproblem,cui2025curieevaluatingllmsmultitask,luo2025bigbenchunifiedbenchmarkevaluating}, yet much of this success may stem from associative recall rather than genuine conceptual reasoning. To probe this distinction, we ask:

\begin{tcolorbox}[colframe=black, colback=white, arc=1mm, boxrule=1pt]
\textbf{Q:} Can LLMs solve AI research problems using \textit{only} their parametric knowledge? 
\end{tcolorbox}

Prior evaluations typically test factual recall, benchmark performance~\citep{srivastava2023beyond,chollet2019measureintelligence}, or domain-specific competence. Our study instead isolates the ability to generate solutions to open-ended research challenges without external aids such as fine-tuning or retrieval augmentation~\citep{Xu2025,zheng2025knowledgeaugmentedcomplexproblem,cui2025curieevaluatingllmsmultitask}.
To that end, we introduce \textsc{AInstein}, a framework for empirically testing scientific problem-solving under these constraints. To create a controlled testbed, we first distill scientific abstracts into concise, solution-free problem statements that preserve the core research challenge while omitting references to the original solution (Section~\ref{sec:preprocessing}). \textsc{AInstein} then tasks solver agents with generating and refining potential solutions through iterative critique loops, following the cycles of proposal, review, and revision characteristic of scientific inquiry. This design separates the problem from the solution, enabling us to distinguish recall from reasoning.

We curate a dataset of 1,214 high-quality ICLR 2025 papers, stratified by quality tier. To ensure reliable evaluation, we first validate our methodology through a blind pairwise study with 20 expert AI researchers (Section~\ref{sec:human_study}), then scale to the full corpus using an LLM-as-a-judge paradigm.

We report two performance metrics for our task: \textbf{Success Rate}: does the proposed solution address the problem statement? \textbf{Rediscovery}: how often does an LLM independently converge on solutions proposed by human researchers?
Together, these metrics allow us to disentangle rote recall from genuine problem-solving. Our evaluation shows a high Success Rate paired with low Rediscovery, implying that models \textit{can} generate novel, valid alternatives rather than reproducing known solutions under certain constraints.

Overall, this work makes three contributions. \textbf{First}, through large-scale evaluation on 1,214 ICLR~2025 papers validated by a blind expert study with 20 AI researchers, we show that LLMs can generate feasible and often novel solutions to research problems from parametric knowledge alone, while revealing that this ability is fragile, particularly when problems require cross-domain transfer. \textbf{Second}, we find that strict rediscovery of published solutions remains rare ($\leq$19\%) despite high success rates ($>$70\%), providing evidence that models engage in genuine problem-solving rather than sophisticated recall. \textbf{Third}, by evaluating on the ResearchPlanGen benchmark~\citep{coscientist2025} across 2,645 problems, we characterize the boundary of what parametric knowledge can achieve through iterative self-critique and revision at inference time (no weight updates) versus what requires task-specific training (e.g., RL fine-tuning).

\section{Related Work}

\paragraph{Evaluating Scientific Knowledge and Reasoning.} Traditional benchmarks have focused on testing models' ability to answer scientific questions, but recent work highlights their conflation of memorization with understanding.~\cite{wang2024scimon} demonstrate that models often rely on pattern matching rather than genuine reasoning. Our work extends this line of inquiry by explicitly controlling for memorization through conceptual abstraction.
Critically, our benchmark draws on ICLR 2025 and 2026 papers, whose acceptance decisions and final manuscripts post-date the training cutoffs of all evaluated models, ensuring that observed performance reflects generalization rather than memorization.

\begin{figure*}[t]
  \centering
  \includegraphics[width=\linewidth]{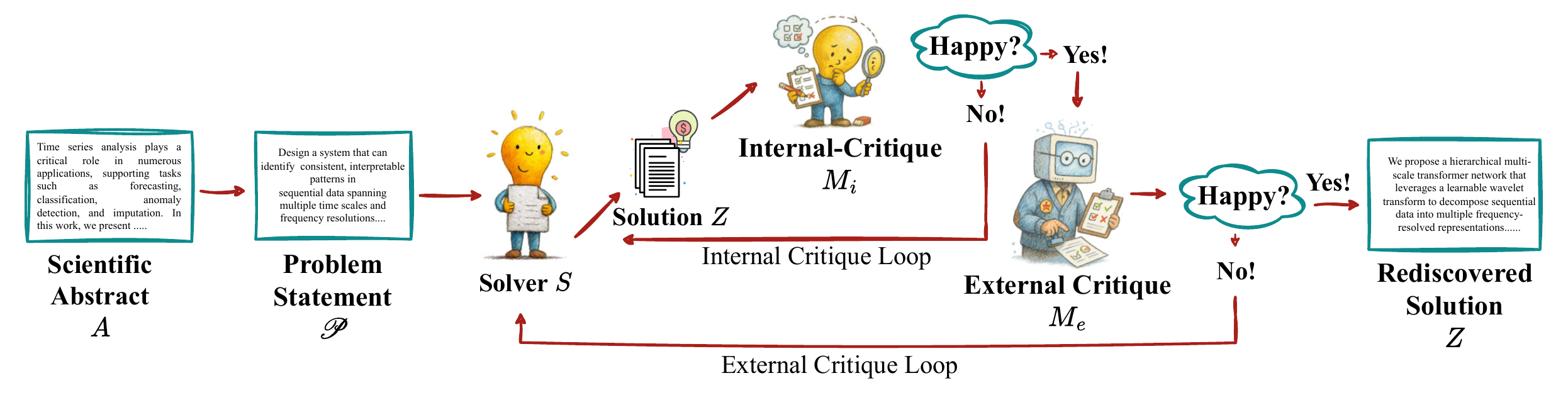}
  \caption{Overview of the evaluation pipeline. A preprocessing Generalizer agent ($\mathcal{G}$) first distills a scientific abstract ($\mathcal{A}$) into a solution-free problem statement ($\mathcal{P}$). The \textsc{AInstein} Solver agent ($\mathcal{S}$) then generates a technical solution ($\mathcal{Z}$) from $\mathcal{P}$ alone. Both stages employ iterative refinement with internal ($\mathcal{M}_i$) and external ($\mathcal{M}_e$) critique (Section~\ref{sec:refinement}).}
  \label{fig:framework}
\end{figure*}


\paragraph{Scientific Discovery with LLM Agents.}
Recent work has pushed LLMs from passive question-answering toward acting as research agents. \citet{romera2024mathematical} use an LLM in an evolutionary loop with an automated evaluator to discover new constructions for the cap-set and bin-packing problems, the first reported case of an LLM-driven open-problem result in mathematics. In the natural sciences, \citet{zhavoronkov2019deep} and \citet{merchant2023scaling} use generative models for molecular and materials design, while \citet{coscientist2025} pair GPT-4 with internet search, code execution, and robotic instrumentation to plan and run chemical experiments autonomously. Building on these foundations, fully agentic ML-research pipelines such as the
AI Scientist~\citep{lu2024aiscientist,yamada2025aiscientistv2} and ResearchAgent~\citep{baek2025researchagent} compose ideation, implementation, evaluation, and writing into closed loops, and agent benchmarks~\citep{chan2024mlebench,huang2024mlagentbench}
score these systems against ML-engineering tasks curated from Kaggle and similar sources.

\paragraph{Research Idea Generation and Novelty.}
A complementary line of work asks whether an LLM can propose a novel idea. \citet{si2025can} run a large-scale blind study with 100+ NLP researchers and find that LLM-generated research ideas are judged more novel than expert ideas but slightly weaker on feasibility, exposing a gap between perceived novelty and practical viability. \citet{wang2024scimon} take this further by optimizing for novelty directly during generation, retrieving inspirations from prior literature and iteratively comparing against them until a sufficient delta is reached.
 
\medskip\noindent
We motivate our setup based on the two limitations identified in prior works. First, idea-generation studies measure novelty against retrieved literature or expert ratings, not against the \emph{solution} the research community actually adopted; an idea judged novel may or may not address the problem the field eventually solved. Second, end-to-end agent benchmarks report success rates on engineering tasks but do not test whether the agent's solution \emph{matches} the published approach. AInstein addresses both gaps by pairing \textbf{Success Rate} (does the proposed solution address the problem) with \textbf{Rediscovery} (does it match the published approach) on ICLR 2025 papers, separating the question of whether a model can solve a research problem from whether it solves it the way the field did.

\section{Methodology}
\label{sec:method}
We present \textsc{AInstein}, a framework for evaluating whether large language models (LLMs) can act as autonomous scientific problem-solvers. The framework tasks solver agents with proposing technically valid solutions to research problems through iterative refinement with nested critique loops, which has been shown to improve LLMs’ problem-solving capability~\citep{shinn2023reflexion,madaan2023self}.
We evaluate models behaviorally, by the solutions they produce, rather than by probing internal representations, as probing techniques~\citep{longpre2021entity,turner2023steering} can reveal what knowledge a model encodes, but cannot tell us whether the model can deploy that knowledge to solve a research problem end-to-end.

\subsection{Problem Preprocessing}
\label{sec:preprocessing}
\label{sec:pipeline}
Before invoking \textsc{AInstein}, we preprocess each scientific abstract into a solution-free problem statement. Given an abstract $\mathcal{A}$, a \textbf{Generalizer} agent $\mathcal{G}$ (an LLM prompted with a specific role) produces a distilled problem statement $\mathcal{P}$. The goal is to create a formulation that is faithful to the original research challenge while being free of solution-specific artifacts. To this end, $\mathcal{G}$ is instructed to maximize \textbf{abstraction} and \textbf{fidelity} while minimizing \textbf{ambiguity} and \textbf{solution leakage} (as defined in our evaluation criteria). This constraint prevents the agent from paraphrasing the solution, ensuring the subsequent solving task is non-trivial. Formally: $\mathcal{P} = \mathcal{G}(\mathcal{A}; \mathcal{M}_i, \mathcal{M}_e)$.
The full Generalizer prompt is in Appendix~\ref{app:prompts} (Prompt~\ref{prompt:generalizer}).
Ground-truth solutions are extracted in a similar fashion from the ICLR papers; we treat the published approach as the canonical solution and use it as the reference against which Rediscovery is measured.

\subsection{The \textsc{AInstein} Solver}
\label{sec:solver}
Given a preprocessed problem statement $\mathcal{P}$, the \textbf{Solver} agent $\mathcal{S}$ proposes a technical solution $\mathcal{Z}$. The Solver has access only to $\mathcal{P}$, not to the original abstract or published solution, and must generate a detailed methodology to address the problem. Formally: $\mathcal{Z} = \mathcal{S}(\mathcal{P}; \mathcal{M}_i, \mathcal{M}_e)$.

\subsection{Iterative Refinement via Critique Loops}
\label{sec:refinement}
The Solver, and during preprocessing the Generalizer, both operate under a nested refinement mechanism that mimics scientific inquiry: iterative proposal, critique, and revision (Algorithm~\ref{algo:ours} in Appendix~\ref{app:algorithm}). Each stage has:
\begin{enumerate}
    \item \textbf{Internal Critique Loop (Model $\mathcal{M}_i$):}  
    The inner loop simulates fast, low-cost self-correction. An initial draft (problem statement or solution) is generated, then reviewed against task-specific criteria by an internal critic (also $\mathcal{M}_i$). The critic scores the draft on task-specific dimensions (1--10 scale) and provides a binary accept/reject judgment extracted from its natural-language assessment. If rejected, the draft is revised. This process repeats for up to \texttt{MaxInternalAttempts} iterations (=20 in our experiments). Prompt~\ref{prompt:generalizer} in Appendix~\ref{app:prompts} shows the generalizer prompt.

    \item \textbf{External Critique Loop (Model $\mathcal{M}_e$):}  
    The outer loop provides a higher-fidelity review from an external model $\mathcal{M}_e$. Once the internal loop converges, $\mathcal{M}_e$ evaluates the candidate artifact. If the external judgment is negative, the feedback from both critics is incorporated into a new attempt, up to \texttt{MaxExternalAttempts} (=20 in our experiments). This ensures that only high-quality outputs are accepted. Prompt~\ref{prompt:solver} in Appendix~\ref{app:prompts} shows the solver prompt.
\end{enumerate}

This dual-loop structure mirrors the dynamics of scientific peer review: rapid local iteration combined with stricter external scrutiny. In practice, convergence is fast: across the ICLR benchmark, 59\% of problems converge in a single external iteration, with a median of 5--6 total LLM calls per problem (P95: $\sim$60). On the ResearchPlanGen benchmark, over 73\% converge in one external iteration (median 3 calls). \\

Several lighter-weight inference-time strategies could in principle be substituted for our internal/external critique design, and we considered each before settling on the present form. \emph{Self-consistency}~\citep{wang2023selfconsistency} samples many independent reasoning paths and majority-votes the final answer; it is highly effective on tasks with a small, discrete answer space, but research-problem solutions are free-form artefacts (technical methodologies) for which ``majority answer'' is not well defined. \emph{Best-of-$N$} with a learned verifier or process reward model~\citep{lightman2024verify} replaces voting with a trained scorer, but constructing a verifier for ML research methodologies would itself require the kind of large-scale, in-distribution supervision our setting deliberately withholds. \emph{Self-Refine}~\citep{madaan2023selfrefine} and \emph{Reflexion}~\citep{shinn2023reflexion} use a single model to critique and revise its own output; recent work shows, however, that intrinsic self-correction is brittle and can even degrade performance when the model is not given an external signal that its current draft is wrong~\citep{huang2024cannotcorrect}. Finally, \emph{multi-agent debate}~\citep{du2023debate} pits symmetric agents against each other; debate is well-suited to factual disagreement, but in our setting the bottleneck is not adjudicating between two confident positions, it is producing a draft that survives a critic with \emph{different} information than the proposer. Task-specific fine-tuning is the natural upper bound on inference-time methods, and we in Section~\ref{sec:rpg} reports a head-to-head comparison and shows that our training-free loop matches the gains of RL fine-tuning on in-distribution problems while remaining model-agnostic.
\paragraph{Design choices.} Our nested loop combines the strengths of these alternatives while avoiding their failure modes. The \emph{internal} critic ($\mathcal{M}_i$) is a cheap model that exploits self-refinement signals to make quick iterations on the solution. The \emph{external} critic ($\mathcal{M}_e$) is a different model acting on the same problem with no access to the in-progress draft's prompt history; this is the asymmetry that prior work on intrinsic self-correction lacks~\citep{huang2024cannotcorrect}. We use binary accept/reject at the external stage (rather than a graded score) to avoid leaking a continuous reward signal that could be optimized against rather than satisfied. The internal loop budgets up to \texttt{MaxInternalAttempts}  revisions before each external review, which keeps total cost low (median 5--6 calls per problem) while preserving the stronger gating signal of an out-of-distribution reviewer. Together, these choices isolate what \emph{parametric knowledge plus inference-time critique} can achieve, without reintroducing any supervised signals.


\section{Experimental Setup}
\label{sec:exp_setup}
We organize the empirical study around three datasets and two metrics. §\ref{sec:human_study} reports results on a held-out ICLR 2026 set (N=20, blind expert evaluation); §\ref{sec:auto_eval} scales to 1,214 ICLR 2025 papers using the LLM-as-a-judge pipeline; and §\ref{sec:rpg} evaluates on the ResearchPlanGen benchmark (2,645 problems).

\subsection{Datasets \& Models}
\label{sec:datasets}

Our experimental dataset consists of 1,214 papers curated from the ICLR 2025 conference submissions, drawing inspiration from the ICLR Dataset \citep{gonzalez2024learning}. This corpus provides a large set of high-quality research problems. Our curated set is intentionally stratified by the papers' final acceptance tiers: \textbf{Oral}, \textbf{Spotlight}, and \textbf{Poster}.
Notably, in our curation process we only include higher-quality samples to analyze how model performance correlates with the perceived quality and impact of the research.
Table~\ref{tab:dataset_distribution} compares the distribution of our final dataset against the full ICLR 2025 corpus.
In addition, we hold out a separate set of \textbf{120 ICLR 2026 submissions} across five domains, exclusively for the expert human study reported in §\ref{sec:human_study}.

To prevent data leakage and ensure our evaluation tests reasoning rather than retrieval, all models used in our experiments have knowledge cutoffs that predate the ICLR 2025 submission deadline.
In addition to our curated ICLR benchmark, we evaluate on \textbf{ResearchPlanGen}~\citep{coscientist2025}, a standardized benchmark comprising 2,645 test problems across ML, arXiv, and biomedical domains, each annotated with expert rubric items and reference solutions (Section~\ref{sec:rpg}).
We experiment with three primary model families selected to represent different scales of capability. The large-scale models include \texttt{GPT-OSS-120B} and \texttt{Qwen-235B}, representing the state-of-the-art in reasoning. The mid-scale model is \texttt{Mistral-24B}, a powerful and widely used alternative.

\begin{wraptable}{r}{0.44\columnwidth}
    \centering
    \scriptsize
    \vspace{-6pt}
    \setlength{\tabcolsep}{2.5pt}
    \renewcommand{\arraystretch}{1.0}
    \begin{tabular}{@{}l rrrr@{}}
    \toprule
    & \multicolumn{2}{c}{\textbf{Full ICLR}} & \multicolumn{2}{c}{\textbf{Curated}} \\
    \cmidrule(lr){2-3} \cmidrule(lr){4-5}
    \textbf{Tier} & \textbf{Cnt} & \textbf{\%} & \textbf{Cnt} & \textbf{\%} \\
    \midrule
    Oral      & 213   & 1.8  & 213 & 17.5 \\
    Spotlight & 379   & 3.2  & 379 & 31.2 \\
    Poster    & 3,111 & 26.7 & 622 & 51.2 \\
    Rejected  & 5,014 & 43.0 & --- & ---  \\
    Withdrawn & 2,946 & 25.3 & --- & ---  \\
    \midrule
    \textbf{Total} & \textbf{11,663} & \textbf{100} & \textbf{1,214} & \textbf{100} \\
    \bottomrule
    \end{tabular}
    \caption{Paper distribution: full ICLR 2025 vs.\ our curated set (N=1,214).}
    \label{tab:dataset_distribution}
    \vspace{-18pt}
\end{wraptable}

Each model can serve as the internal model ($\mathcal{M}_i$, which generates drafts) or the external model ($\mathcal{M}_e$, which provides critique). We systematically test all pairings to analyze the interplay between generation and critique capabilities, avoiding biases that might arise from using models of the same family in both roles.

\subsection{Evaluation}

\paragraph{Expert Human Validation.} We conduct a blind pairwise study with 20 expert AI researchers across five specializations. Critically, the study uses held-out ICLR 2026 problems to avoid data contamination and ensure that LLM solutions reflect reasoning rather than memorized content. Each expert evaluated 8 problems across three dimensions: \textit{Feasibility} (is the proposed approach technically sound and feasible to implement?), \textit{Solvability} (would the proposed approach plausibly solve the stated problem?), and \textit{Originality} (is the proposed approach distinct from existing work in the problem space?). The full judge prompt is in Appendix~\ref{fig:study_welcome} and more study design details are included in Appendix~\ref{app:study_design}.

\paragraph{Automated Evaluation.} We use GPT-OSS-120B as an LLM-as-a-judge~\citep{liu2023g,sottana2023evaluation}, which is shown to correlate well with human evaluation in our study.
For each generated problem statement (P), we calculate a deficit score $d \in [0, 10]$ (lower is better) as the average penalty across four dimensions: \textit{Fidelity}, \textit{Abstraction}, \textit{Ambiguity}, and solution \textit{Leakage}. For each solution (Z), the LLM judge scores technical \textit{Feasibility}, \textit{Solvability}, and \textit{Originality} on a 1--5 scale. These scores inform two primary metrics: \textbf{Success Rate} (is the solution both feasible and complete, i.e., score $\geq 4$?) and \textbf{Rediscovery} (does a successful solution match the original human solution?). Additional quantitative text analyses are described in Appendix~\ref{app:quant_metrics}.

\section{Results}
\label{sec:results}


\subsection{Expert Human Validation}
\label{sec:human_study}
We begin with the most direct form of evidence: expert assessment. On the held-out \textbf{ICLR 2026} set introduced in §\ref{sec:datasets}, we conduct a targeted human study in which expert AI researchers perform blind pairwise comparisons between LLM-generated and published solutions~\citep{gu2024survey,zheng2023judging}. Because every ICLR 2026 problem post-dates the training cutoff of all evaluated models, this section isolates the contribution of parametric knowledge from any possibility of memorized exposure.

\paragraph{Study Design.} We recruited $N\!=\!20$ AI researchers (PhD students, postdocs, and faculty) with at least two years of research experience, balanced at 4 per expertise category across five specializations. Critically, this study uses problems drawn from ICLR 2026 submissions to ensure that LLM solutions reflect reasoning rather than memorized content. Each researcher evaluated 8 problems, viewing two anonymized solutions side-by-side and indicating a preference (A, B, or Tie) on three dimensions: \textbf{Feasibility of Implementation}, \textbf{Solves the Problem}, and \textbf{Originality}. Solution labels were randomly flipped to control for position bias~\citep{wang2024pairwise}, and shared anchor problems enabled inter-rater reliability analysis. Using $\mathcal{M}_i\!=\!$GPT-OSS-120B throughout, the study yielded 480 ratings across 96 problems in three pair types: LLM vs.\ Human (two $\mathcal{M}_e$ variants) and GPT vs.\ Mistral (full details in Appendix~\ref{app:study_design}).

\begin{table}[t]
\centering
\small
\setlength{\tabcolsep}{3.5pt}
\begin{tabular}{l cc cc cc}
\toprule
& \multicolumn{2}{c}{\textbf{Feasibility}}
& \multicolumn{2}{c}{\textbf{Solves Problem}}
& \multicolumn{2}{c}{\textbf{Originality}} \\
\cmidrule(lr){2-3}\cmidrule(lr){4-5}\cmidrule(lr){6-7}
\textbf{Comparison ($\mathcal{M}_e$)} & Win\% & $p$ & Win\% & $p$ & Win\% & $p$ \\
\midrule
GPT-OSS-120B vs.\ Human  & 48.9 & 1.00  & \textbf{76.2} & 9.4e-4 & \textbf{72.1} & .005 \\
Mistral-24B vs.\ Human   & 56.0 & .48   & \textbf{70.7} & .01    & 58.7          & .30  \\
Combined LLM vs.\ Human  & 52.6 & .68   & \textbf{73.5} & 2.2e-5 & \textbf{65.2} & .006 \\
\midrule
GPT-OSS-120B vs.\ Mistral-24B & 50.0 & 1.00 & 45.5 & .73 & 67.9 & .09 \\
\bottomrule
\end{tabular}
\caption{Expert human evaluation ($N\!=\!20$, ICLR 2026 problems).
Win\,\% excludes ties; $p$ from two-sided binomial test.
$\mathcal{M}_i\!=\!$GPT-OSS-120B throughout.}
\label{tab:human_eval}
\end{table}

\begin{figure*}[t]
    \centering
\includegraphics[width=0.95\textwidth]{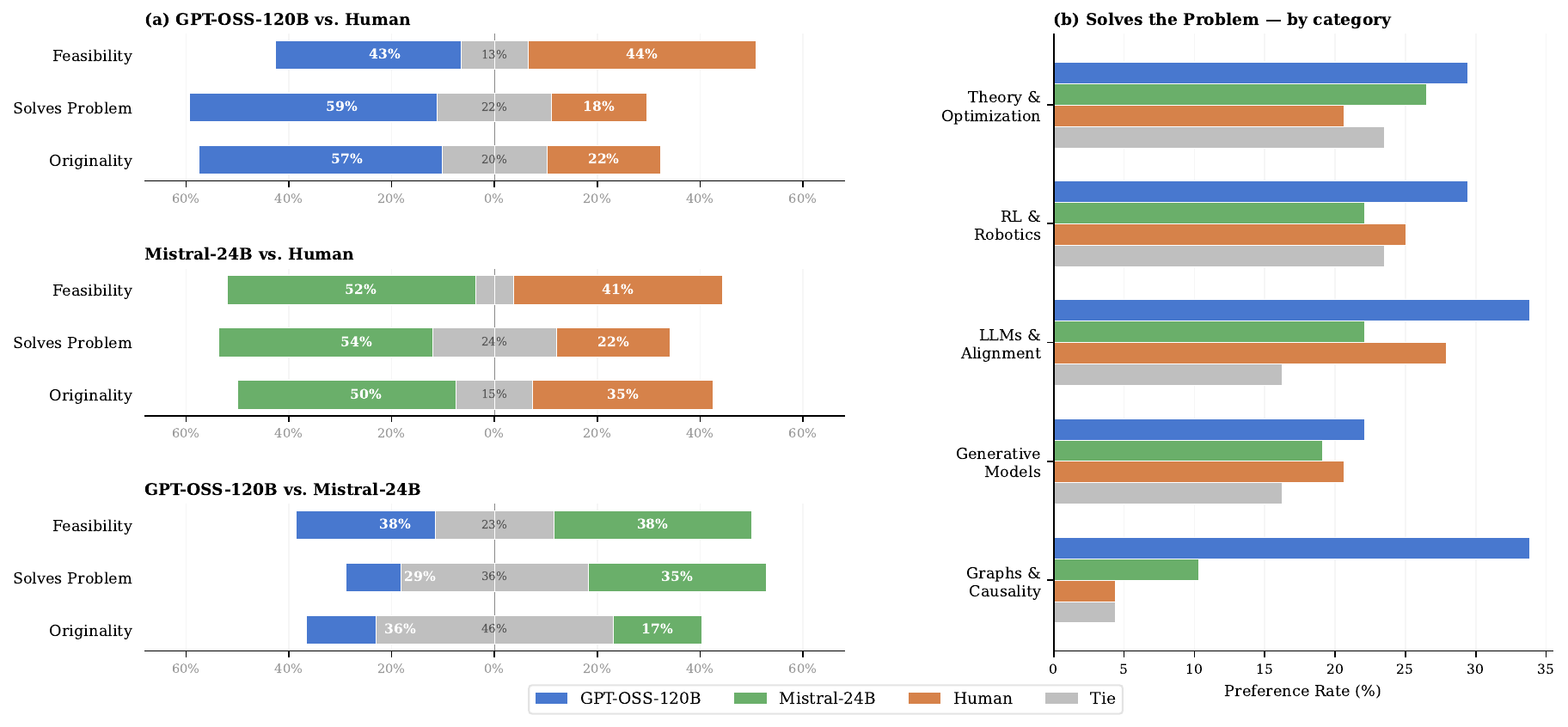}
    \caption{Expert human evaluation ($N\!=\!20$, ICLR~2026). \textbf{(a)}~Pairwise preference rates: LLMs dominate on \emph{Solves the Problem} and \emph{Originality}; Feasibility shows no significant difference. \textbf{(b)}~The LLM advantage is consistent across all five expertise categories.}
    \label{fig:human_eval}
\end{figure*}

\paragraph{Finding 1: LLM solutions are competitive on feasibility
and significantly preferred on problem-solving and originality.}
Table~\ref{tab:human_eval} and Figure~\ref{fig:human_eval}
summarize the pairwise preferences.
On \emph{Solves the Problem}, LLM solutions are preferred in
\textbf{73.5\%} of decided comparisons
($p\!=\!2.2{\times}10^{-5}$) and on \emph{Originality}, they are preferred in
\textbf{65.2\%} ($p\!=\!.006$).
\emph{Feasibility} shows no significant difference
($p\!=\!.685$), indicating that LLM-generated proposals
achieve comparable implementation feasibility to published work
while exploring a broader solution space. This suggests that parametric knowledge, when elicited through structured critique, may encode sufficient methodological understanding to produce implementable research proposals. These results are consistent with \citet{si2025can}'s finding
that AI-generated ideas receive higher novelty ratings than
expert ideas, and extend that observation from open-ended
ideation to technically detailed solution generation.
The effect is robust across all five expertise categories
(Figure~\ref{fig:human_eval}b), ruling out domain-specific
confounds, and strengthens among high-confidence ratings
($\geq\!4$): preference rises to 77.4\% ($p\!<\!10^{-4}$).

\paragraph{Finding 2: GPT-OSS-120B outperforms Mistral-24B
when evaluated against human baselines.}
Further decomposing our results by model reveals an asymmetry in the strength of the LLM advantage. GPT-OSS-120B achieves significant wins
on both \emph{Solves the Problem} (76.2\%,
$p\!=\!9.4{\times}10^{-4}$) and \emph{Originality}
(72.1\%, $p\!=\!.005$), whereas Mistral-24B reaches
significance only on problem-solving (70.7\%, $p\!=\!.01$)
but not on Originality ($p\!=\!.30$).
This difference suggests that increased model capacity
contributes to more creative problem-solving, beyond
improvements in correctness alone.

\paragraph{Finding 3: GPT-OSS-120B and Mistral-24B are
statistically indistinguishable in direct comparison.}
In head-to-head evaluation (Table~\ref{tab:human_eval},
bottom row), the two models show no significant difference on any dimension (all $p\!>\!.09$), accompanied by high tie rates of 23--46\% (Figure~\ref{fig:human_eval}a, bottom panel).
GPT-OSS-120B exhibits a marginal advantage on Originality
(67.9\% win rate, $p\!=\!.09$). 

\paragraph{Inter-rater agreement.}
Inter-rater agreement on 20 shared anchor problems yields Fleiss'~$\kappa\!=\!0.24$ overall (0.28 for Feasibility, 0.16 for Solves the Problem, 0.28 for Originality). Pairwise agreement is 63.5\%, with 43.8\% of pairs in perfect agreement, indicating moderately fair reliability~\citep{landis1977measurement}. This is consistent with the known difficulty of evaluating open-ended scientific proposals.

\subsection{The Spectrum of Parametric Problem-Solving}
\label{sec:spectrum}

The expert study establishes that LLM solutions are competitive with published research. But what does parametric problem-solving look like in practice?

We show three representative examples illustrate the spectrum (full solution texts in Appendix~\ref{app:case_studies}).
\textbf{Successful rediscovery.} Given a problem requiring associative memory for hierarchical data, both GPT-OSS-120B and Mistral-24B independently converge on the central insight of the published solution: extending Hopfield retrieval to hyperbolic space. GPT-OSS-120B proposes a Lorentz-manifold formulation with Riemannian optimization; Mistral-24B proposes a Poincar\'e-ball variant with M\"obius layers. That two independently prompted models recover the same non-trivial geometric construction illustrates what Rediscovery captures.
\textbf{Valid but conventional.} For a problem requiring lightweight GNN calibration with OOD robustness, the published paper adopts a unique cross-domain approach via spiking neural networks and predictive coding. Both LLMs instead propose established probabilistic methods (Bayesian message passing and Dirichlet Prior Networks) that are technically sound but follow well-explored directions.
\textbf{Missed cross-domain transfer.} On a problem requiring acceleration of sequential AI agent execution, the published solution draws on CPU speculative execution. Both LLMs propose implementation-level optimizations (result memoization, ONNX compilation) that address the symptom but do not recover the architectural insight.

These cases reveal a consistent pattern: LLMs reliably reconstruct solutions within familiar methodological boundaries but struggle with cross-domain analogical transfer. A systematic clustering of the full solution space (KMeans++ on \texttt{Qwen-8B} embeddings) corroborates this pattern, revealing 11 distinct solution archetypes (Table~\ref{tab:cluster_archetypes} and Appendix~\ref{app:visual_clusters}--\ref{app:cluster_extend}).

\subsection{Automated Evaluation at Scale}
\label{sec:auto_eval}

We now report results on our LLM-as-a-judge evaluation on 1,214 ICLR 2025 papers, confirming the expert findings while revealing finer-grained distinctions across model configurations and paper tiers.

\paragraph{Preprocessing Quality.}
The validity of the automated study depends on the quality of the preprocessed problem statements provided to the Solver. As shown in Table~\ref{tab:gen_metrics} (Appendix~\ref{app:gen_metrics}), both GPT-OSS-120B and Qwen-235B produce problem statements with low deficit scores ($d \approx 2.5$), while Mistral-24B yields moderately lower quality ($d \approx 3.5$). The deficit score correlates strongly with Information Loss and Ambiguity (Figure~\ref{fig:gen_corr_heatmap} in Appendix~\ref{app:gen_corr}), and statistical testing confirms that GPT-OSS-120B and Qwen-235B perform comparably (Appendix~\ref{app:sig_testing}). We evaluate Solver agents on problems from all three Generalizers to ensure robustness to variation in problem framing.

\paragraph{The Internal Model's Capability is Paramount.} The automated results (Table~\ref{tab:solver_results}) reveal a clear performance hierarchy consistent with the human evaluation. \texttt{GPT-OSS-120B} as the internal model achieves a Success Rate of 70--79\% across all model configurations (Appendix~\ref{app:full_solver}), substantially outperforming \texttt{Qwen-235B} (39--51\%) and \texttt{Mistral-24B} (31--44\%). This gap is larger than what human experts perceive in pairwise comparison, suggesting that the automated judge captures performance differences that are difficult to detect in a head-to-head format.

\paragraph{Rediscovery vs.\ Creative Problem-Solving.} Rediscovery rates are notably low (14--19\% even for the best agent), confirming that perfect reproduction of human solutions is exceptionally rare. Combined with high Success Rates ($>$70\%), this gap implies that the majority of valid solutions are novel alternatives rather than rediscoveries, a finding directly corroborated by the human study, where experts rated LLM solutions as significantly more original than published abstracts.

\begin{table}[t]
\centering
\scriptsize
\setlength{\tabcolsep}{4pt}
\renewcommand{\arraystretch}{1.0}
\begin{tabular}{@{}ll cc@{}}
\toprule
\multicolumn{2}{c}{\textbf{Model Configuration}} & \textbf{Rediscovery} & \textbf{SR Solver} \\
\cmidrule(r){1-2} \cmidrule(l){3-4}
\textbf{External Model} & \textbf{Internal Model} & & \\
\midrule
\multirow{3}{*}{GPT-OSS-120B}
  & \textbf{GPT-OSS-120B} & \textbf{19.11} & \textbf{74.05} \\
  & Qwen-235B             & 7.74  & 43.82 \\
  & Mistral-24B           & 6.43  & 34.60 \\
\addlinespace[2pt]
\midrule
\multirow{3}{*}{Mistral-24B}
  & \textbf{GPT-OSS-120B} & \textbf{17.79} & \textbf{71.91} \\
  & Qwen-235B             & 7.17  & 39.21 \\
  & Mistral-24B           & 6.75  & 31.22 \\
\addlinespace[2pt]
\midrule
\multirow{3}{*}{Qwen-235B}
  & \textbf{GPT-OSS-120B} & \textbf{17.38} & \textbf{70.35} \\
  & Qwen-235B             & 7.66  & 42.34 \\
  & Mistral-24B           & 7.08  & 34.84 \\
\bottomrule
\end{tabular}
\caption{Solver performance (\%) with a Qwen-235B as LLM judge. Bold = best
$\mathcal{M}_i$ per metric. Results are consistent across
problem sources (Appendix~\ref{app:full_solver}).}
\label{tab:solver_results}
\end{table}

\paragraph{Performance Across Paper Tiers.} Our stratification by paper quality (Oral, Spotlight, Poster) reveals that the performance hierarchy remains remarkably consistent across all three tiers (Figure~\ref{fig:performance_by_tier} in Appendix~\ref{app:paper_tiers}). The top-performing \texttt{GPT-OSS-120B} agent shows stable Success Rates for Oral (69.0\%), Spotlight (77.8\%), and Poster (72.5\%) papers. This result suggests that the model's ability to generate a valid solution is not significantly hindered by the novelty or the perceived impact of the original paper.

\paragraph{Robustness to Evaluator Choice.} To verify that our findings are not an artifact of a specific judge, we replaced \texttt{GPT-OSS-120B} with \texttt{Qwen3-235B} and replicated the full evaluation. The same performance hierarchy holds: \texttt{GPT-OSS-120B} as $\mathcal{M}_i$ achieves 84.1\% SR Solver (vs.\ 74.1\% with the GPT judge) and 17.9\% Rediscovery (vs.\ 19.1\%), preserving its dominance over \texttt{Qwen-235B} (67.0\%/7.8\%) and \texttt{Mistral-24B} (50.5\%/6.4\%). The full cross-model evaluation is in Appendix~\ref{app:qwen_eval}. Combined with the independent expert validation (Section~\ref{sec:human_study}), this provides strong evidence that our findings are robust to evaluator choice.

\paragraph{Convergence Across Modalities.}
The human and automated evaluations, conducted on separate corpora (ICLR~2026 and 2025 respectively) using independent assessment methods, converge on three conclusions. \textbf{(1)}~LLM solutions are competitive with published research: experts prefer them on problem-solving ($p\!=\!2.2{\times}10^{-5}$) and originality ($p\!=\!.006$), with no significant difference on feasibility. \textbf{(2)}~The internal model is the dominant factor: $\mathcal{M}_e$ variation accounts for at most 4~percentage points, and both modalities place GPT-OSS-120B as $\mathcal{M}_i$ at the top. \textbf{(3)}~Models generate novel solutions rather than reproducing established ones: strict Rediscovery remains rare ($\leq$19\%) despite high Success Rates ($>$70\%), and experts independently rate LLM solutions as significantly more original. That these conclusions hold across independent corpora, evaluation methods, and benchmarks provides a degree of triangulation uncommon in LLM evaluation studies.

\subsection{Evaluation on the ResearchPlanGen Benchmark}
\label{sec:rpg}

The preceding sections establish what LLMs can solve from parametric knowledge on a curated ICLR benchmark. A natural follow-up is where the limits of training-free parametric approaches lie. We evaluate on \textbf{ResearchPlanGen}~\citep{coscientist2025}, a standardized benchmark for AI-generated research plans. \citet{coscientist2025} showed that RL-finetuned models can achieve 12--22\% relative improvement over base models on rubric satisfaction scores by training on thousands of goal-specific grading rubrics. A key question is whether \textsc{AInstein}'s test-time iterative refinement, which requires \textit{no task-specific training}, or \textit{internet search} can achieve comparable gains purely through inference-time compute.

The benchmark defines 2,645 test problems across three domains (machine learning, 685 items; general arXiv, 1,496; biomedical research, 464), each paired with 10 expert-written rubric items and reference solutions.

\paragraph{Evaluation Protocol.}
Following~\citet{coscientist2025}, we evaluate rubric satisfaction using the 7-desiderata framework: for each rubric item, an LLM judge (\texttt{GPT-OSS-120B}) assesses whether the generated plan satisfies seven quality desiderata (handles all criteria, detailed implementation, no overlooked flaws, well-justified rationale, cost-efficient, no ethical issues, internally consistent). A rubric item is satisfied only if no desiderata are violated. The final score is the fraction of rubric items fully satisfied.

\begin{figure}[t]
    \centering
    \includegraphics[width=\columnwidth]{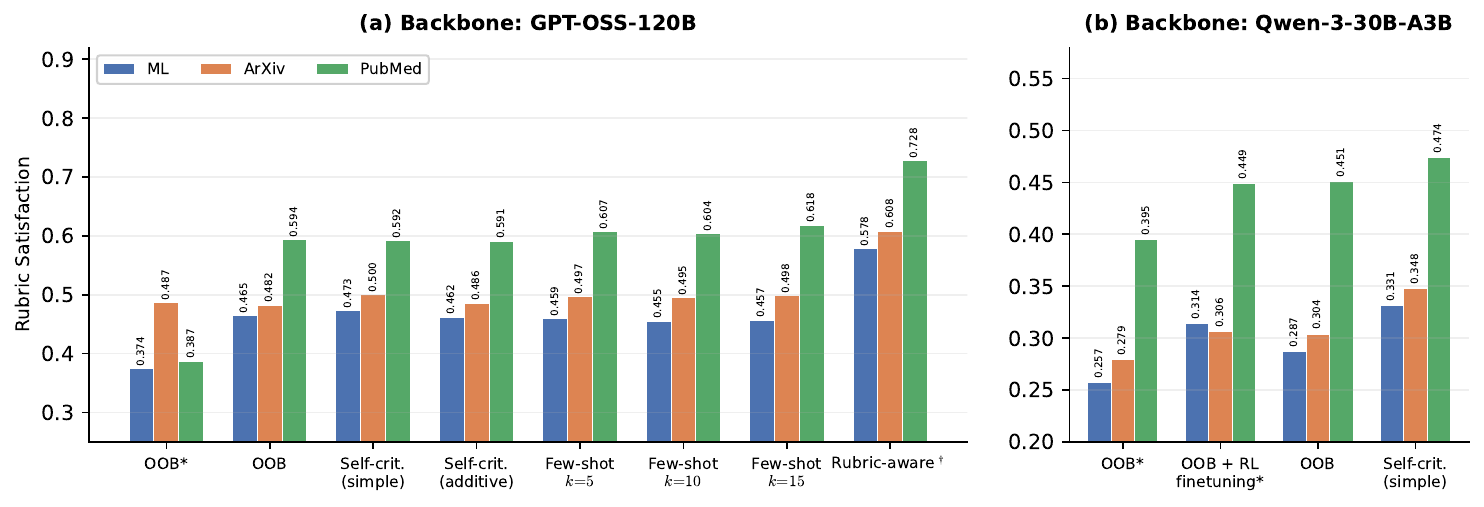}
    \caption{Rubric satisfaction on ResearchPlanGen (7-desiderata evaluation). OOB = out-of-the-box; * = proprietary judge from \citet{coscientist2025}; unmarked = our \texttt{GPT-OSS-120B} judge; $\dagger$ = oracle with ground-truth rubric. \textbf{(a)}~Rubric-aware refinement yields large gains; training-free methods improve modestly. \textbf{(b)}~With Qwen-3-30B-A3B, training-free self-criteria surpasses the RL-finetuned model from \citet{coscientist2025} on all three domains.}
    \label{fig:rpg_results}
\end{figure}

\paragraph{Experimental Conditions.}
All conditions use \texttt{GPT-OSS-120B} as both $\mathcal{M}_i$ and $\mathcal{M}_e$. The \textbf{out-of-the-box} baseline is single-pass generation with no refinement. \textbf{AInstein (self-criteria)} runs the full refinement loop with the reviewer self-generating evaluation criteria without seeing the test rubric, in two variants: \textit{simple} (``identify 5--8 criteria'') and \textit{additive} (iteratively build criteria from multiple facets). \textbf{AInstein (few-shot)} adds $k$ ({=}5, 10, 15) training examples with rubrics as in-context demonstrations. Finally, the \textbf{rubric-aware} oracle ($\dagger$ in Figure~\ref{fig:rpg_results}) provides the actual test rubric items to solver and reviewer.

\paragraph{Results.} Figure~\ref{fig:rpg_results} presents rubric satisfaction scores across all three domains. Four key findings emerge:

\emph{(1) Evaluator choice substantially affects absolute scores.} Comparing the out-of-the-box rows reveals that the \citet{coscientist2025} judge (a proprietary ensemble of GPT-5, Claude, and Gemini) and our \texttt{GPT-OSS-120B} judge can diverge considerably: on PubMed our judge scores substantially higher (0.594 vs.\ 0.387), while on ArXiv scores are comparable (0.482 vs.\ 0.487).
Since we lack access to proprietary models, all subsequent comparisons use our judge for internal consistency (which also showed alignment with human judgements from our study).

\emph{(2) Rubric-aligned refinement is highly effective.} The rubric-aware oracle achieves substantial gains over the out-of-the-box baseline (+24.3\% on ML, +26.1\% on ArXiv, +22.6\% on PubMed), demonstrating that \textsc{AInstein}'s iterative refinement mechanism produces large improvements when the reviewer's criteria are aligned with the evaluation metric.

\emph{(3) Training-free refinement yields consistent gains, competitive with RL finetuning.} For \texttt{GPT-OSS-120B}, simple self-criteria and few-shot $k$=15 outperform the out-of-the-box baseline by 2--4\% across domains, with the largest gains on ArXiv (+3.7\%, self-criteria) and PubMed (+4.0\%, few-shot $k$=15). Notably, for \texttt{Qwen-3-30B-A3B}, \textsc{AInstein}'s training-free self-criteria refinement outperforms the RL-finetuned model from \citet{coscientist2025} on all three domains (0.331 vs.\ 0.314 on ML, 0.348 vs.\ 0.306 on ArXiv, 0.474 vs.\ 0.449 on PubMed), demonstrating that iterative test-time refinement can match or exceed task-specific training for smaller models.

\emph{(4) Criteria alignment is the ceiling.} A substantial gap remains between the best generic method and the rubric-aware oracle. A manual coverage analysis on the ML split reveals that self-generated criteria semantically cover approximately 70\% of ground-truth rubric items; the missed $\sim$30\% are predominantly problem-specific requirements (named techniques, domain constraints, benchmark-specific validation) that cannot be reliably inferred from the problem statement alone without task-specific training~\citep{coscientist2025} or rubric access. This finding characterizes a concrete boundary of parametric problem-solving: iterative test-time refinement helps, but its ceiling is determined by how well the reviewer's self-generated criteria align with the evaluation metric.

\section{Discussion}
\label{sec:discussion}

\paragraph{A parametric knowledge boundary.}
Our results reveal a consistent \textit{parametric knowledge boundary}: LLMs reconstruct solutions within familiar methodological territories (e.g., extending Hopfield retrieval to hyperbolic space, applying standard probabilistic calibration) but fail when solutions require analogical transfer across domain boundaries (e.g., mapping CPU speculative execution onto AI agent acceleration). This pattern held across all three evaluation modalities and model scales, suggesting that parametric training encodes methods and their established domains of applicability, but not the cross-domain bridges that enable the most creative human solutions. The ResearchPlanGen criteria-coverage analysis independently corroborates this boundary: self-generated criteria cover approximately 70\% of ground-truth rubric items, with the missed $\sim$30\% consisting predominantly of problem-specific requirements that cannot be inferred from general methodological knowledge alone.

\paragraph{Genuine exploration, not sophisticated recall.}
Perhaps the most striking finding is the combination of high Success Rates ($>$70\%) with low Rediscovery ($\leq$19\%). Models do not merely reproduce known solutions; they generate valid \textit{alternatives} that domain experts independently rate as significantly more original than published approaches ($p\!=\!.006$). This suggests that LLMs have internalized a broad repertoire of methodological primitives from training and can recombine them into coherent proposals. It also implies that the solution space for most research problems is wider than any single published approach suggests. This reframes the evaluation question: rather than asking whether LLMs can match human output, the more productive question is whether they can systematically explore the space of valid alternatives, and where that exploration breaks down.

\paragraph{Implications for AI-assisted research.}
The parametric knowledge boundary suggests a natural division of labor: LLMs as efficient generators of in-domain methodological proposals, with human researchers providing the cross-domain analogical bridges that current models cannot. The iterative refinement mechanism is the key differentiator over raw generation; it transforms latent parametric knowledge into structured, critique-tested proposals. That 59\% of problems converge in a single external iteration (median 5--6 LLM calls) indicates this process is surprisingly efficient. The RPG finding that training-free refinement matches RL finetuning for smaller models (the Qwen-3-30B comparison) carries practical implications: structured inference-time compute may offer a viable alternative to expensive task-specific training pipelines for research assistance tools.

\paragraph{Future directions.}
The parametric knowledge boundary is a concrete, measurable target. Retrieval-augmented generation, tool use, or explicit training on cross-domain analogies are candidate mechanisms for extending what parametric knowledge alone cannot reach. \textsc{AInstein}'s modular design, separating preprocessing from solution generation, allows future work to plug in retrieval- or tool-augmented solvers while retaining the same evaluation harness. The combination of Success Rate and Rediscovery directly measures whether new approaches expand the boundary of what models can solve versus what they can only recall. We release the full evaluation pipeline, the curated 1,214-problem ICLR benchmark, and all model outputs to serve as a testbed for future work on scientific reasoning in LLMs.

\section*{Limitations}

We note the following limitations in the current study. First, the solutions produced by \textsc{AInstein} are conceptual proposals that have not been implemented or experimentally validated; demonstrating that an approach is feasible on paper does not guarantee it works in practice. Second, although we use ICLR 2025 papers for benchmarking, it is possible that some were publicly available before the training cutoffs of the models we evaluate (like Jun 2024 for GPT-OSS-120B), raising the possibility that some solutions reflect memorized content rather than genuine reasoning. Our ICLR 2026 held-out set mitigates this for the human study. Third, although the nested refinement loop permits up to $20 \times 20$ iterations, the actual cost is much lower in practice (median 5--6 LLM calls on ICLR, 3 on ResearchPlanGen); nevertheless, worst-case runs can reach $\sim$200 calls. Fourth, the human validation study is modest in scale ($N\!=\!20$ experts, 120 problems) with fair inter-rater reliability ($\kappa\!=\!0.24$). Finally, while we frame our evaluation as testing ``parametric knowledge,'' the framework involves elaborate multi-agent prompt engineering and iterative refinement; the boundary between what the model ``knows'' and what the framework ``elicits'' warrants further investigation.

\section*{Ethics Statement}

This work evaluates LLMs' ability to generate conceptual solutions to research problems; no executable code, experimental systems, or real-world deployments are produced. Our human evaluation study involved 20 expert AI researchers who participated voluntarily with informed consent. All evaluations were conducted on anonymized problem-solution pairs derived from publicly available conference submissions (ICLR 2025 and 2026), and no personally identifiable information was collected or stored.

We acknowledge the dual-use potential of AI systems capable of generating research proposals. While \textsc{AInstein} is designed as an evaluation framework for studying LLM capabilities, the underlying methodology could in principle be repurposed to generate misleading or fabricated research content. We believe that openly studying and characterizing these capabilities, including their limitations, contributes to the responsible development of AI for science by enabling the community to develop appropriate safeguards. Our finding that LLM-generated solutions are primarily conceptual and have not been experimentally validated (as noted in Limitations) provides an important caveat against uncritical adoption of AI-generated research outputs.
All models used in this study are publicly available or accessible through standard APIs. Our benchmark datasets are derived from publicly available conference proceedings and the open ResearchPlanGen benchmark~\citep{coscientist2025}.

\bibliography{colm2026_conference}
\bibliographystyle{colm2026_conference}

\appendix
\section*{Appendix}
\appendix

\section{LLM Usage}
We have used LLMs to polish the text of the paper, particularly, to enhance the flow. We also chatGPT's deep research feature to retrieve some relevant papers in the landscape on top on the foundational papers we knew of.

\section{Human Study Design Details}
\label{app:study_design}

The 20 expert AI researchers were balanced at 4 per expertise category: LLMs \& Alignment, Generative Models \& Vision, RL \& Robotics, Theory \& Optimization, and Graphs, Causality \& Structured Reasoning. All had at least two years of active AI/ML research experience and prior peer-review experience. We applied the same problem geneartion and Solver pipeline (Section~\ref{sec:pipeline}) to 120 ICLR 2026 abstracts stratified across tiers, from which the study problems were sampled. Of the 8 problems assigned to each researcher, 4 were shared anchor problems within each category for inter-rater reliability analysis. The study yielded 480 total ratings across 96 problems and three pair types: Human vs.\ $\mathcal{M}_e =$ GPT-OSS-120B ($n\!=\!162$) and Human vs.\ $\mathcal{M}_e =$ Mistral-24B ($n\!=\!162$) test whether LLM solutions are competitive with published research, while GPT vs.\ Mistral ($n\!=\!156$) tests whether the automated judge's model ranking is confirmed by experts.

\subsection{Human Study Interface and Instructions}
\label{app:study_interface}

Participants accessed the evaluation platform at a
dedicated web application. Upon login, each researcher was
shown a welcome page with the full study instructions,
including definitions for all three evaluation dimensions. A practice problem was
provided to familiarize participants with the interface
before recorded evaluations began.

\begin{figure*}[h]
    \centering
    \includegraphics[width=0.95\textwidth]{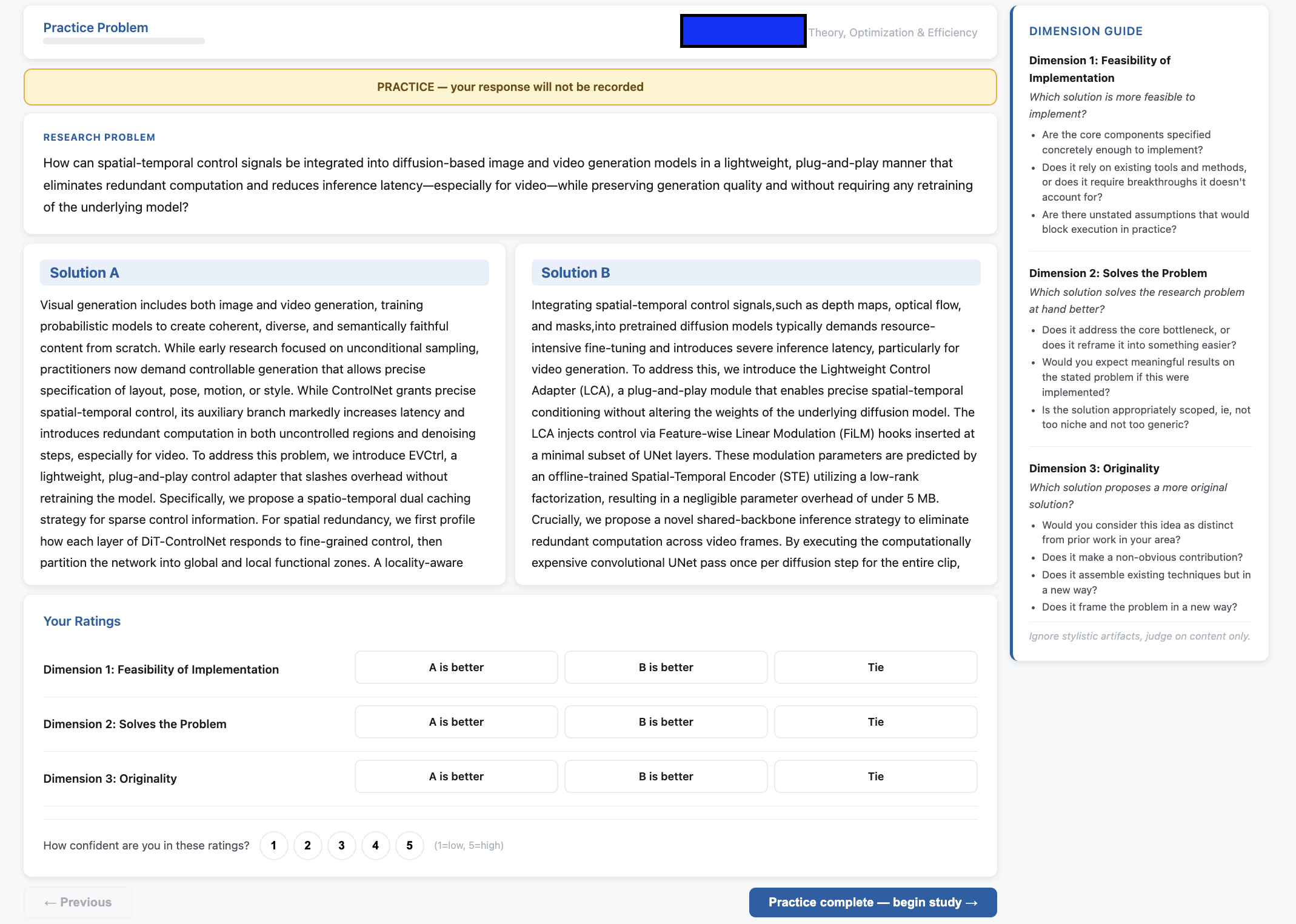}
    \caption{Evaluation interface for a single problem
    (practice round shown). The research problem appears at
    the top, two anonymized solutions are displayed
    side-by-side, and the dimension guide is accessible in
    the right panel. Participants rate each dimension and
    provide a confidence score (1-5).}
    \label{fig:study_interface}
\end{figure*}

Figure~\ref{fig:study_interface} shows the evaluation
interface for a single problem. Each screen presents a
research problem at the top, followed by two anonymized
solutions (A and B) displayed side-by-side. The right panel
contains the dimension guide for reference throughout the
evaluation. Below the solutions, participants indicated
their preference (A is better, B is better, or Tie) on
each of the three dimensions and provided a confidence
rating (1-5). Solution assignment to positions A and B was
randomized per researcher-problem pair to control for
position bias.

\begin{figure}[h]
    \centering
    \includegraphics[width=\columnwidth]{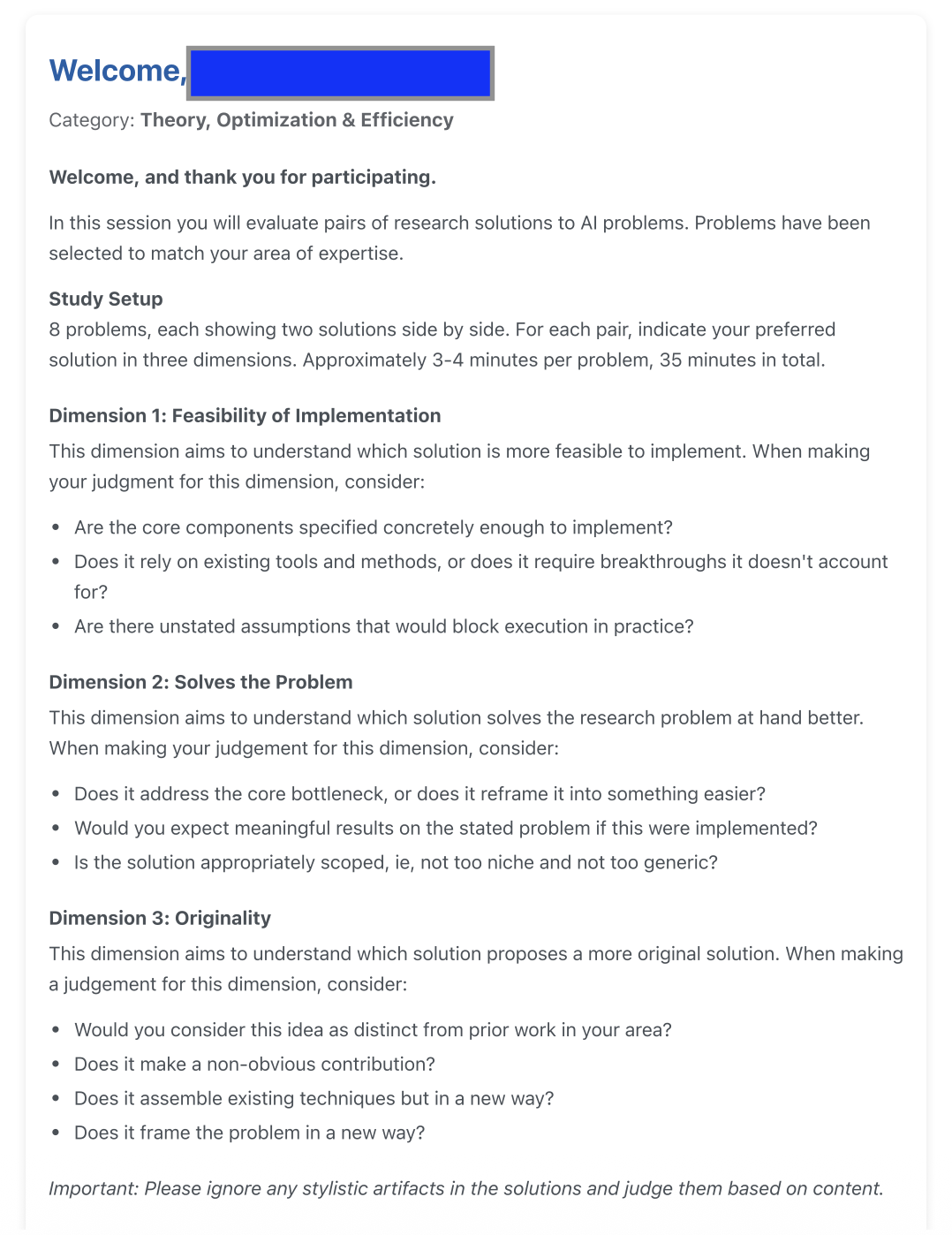}
    \caption{Welcome page and instructions shown to each
    participant upon login. The three evaluation dimensions
    and their guiding criteria are defined before the study
    begins.}
    \label{fig:study_welcome}
\end{figure}

\section{Algorithm}
\label{app:algorithm}
Algorithm~\ref{algo:ours} formalizes the nested
internal-external critique loop used by both the
Generalizer and Solver agents (\S\ref{sec:pipeline}).

\begin{algorithm}[h]
\caption{The AInstein Algorithm}
\label{algo:ours}
\begin{algorithmic}[1]
\State \textbf{Input:} Abstract $\mathcal{A}$, Internal model $\mathcal{M}_i$, External model $\mathcal{M}_e$
\State \textbf{Output:} Problem $\mathcal{P}_{final}$, Solution $\mathcal{Z}_{final}$
\State $\mathcal{P}_{final}, \mathcal{P}_{candidate} \gets \text{null}$
\For{$e=1$ \textbf{to} MaxExternalAttempts}
  \For{$i=1$ \textbf{to} MaxInternalAttempts}
    \State Generate $\mathcal{P}_{candidate}$; self-critique($\mathcal{P}_{candidate}$)
    \If{pass} \textbf{break}
    \EndIf
  \EndFor
  \State External critique($\mathcal{P}_{candidate}$)
  \If{pass} $\mathcal{P}_{final} \gets \mathcal{P}_{candidate}$; \textbf{break}
  \EndIf
\EndFor
\If{$\mathcal{P}_{final} \neq \text{null}$}
  \State $\mathcal{Z}_{final} \gets$ SolveWithRefinement($\mathcal{P}_{final}, \mathcal{M}_i, \mathcal{M}_e$)
\EndIf
\end{algorithmic}
\end{algorithm}

\section{Quantitative Validation Metrics}
\label{app:quant_metrics}

For semantic coherence, we generate 4096-dimensional embeddings using \texttt{Qwen3-Embedding-8B}. To optimize for the target relationship (e.g., problem-solution alignment vs. end-to-end rediscovery), we prepend each text pair with a task-specific instruction before embedding. From these embeddings, we assess textual complexity using standard readability scores (e.g., Flesch-Kincaid Grade Level).

\section{Generalizer Correlation Analysis}
\label{app:gen_corr}
\begin{figure}[h]
    \centering
    \includegraphics[width=0.5\textwidth]{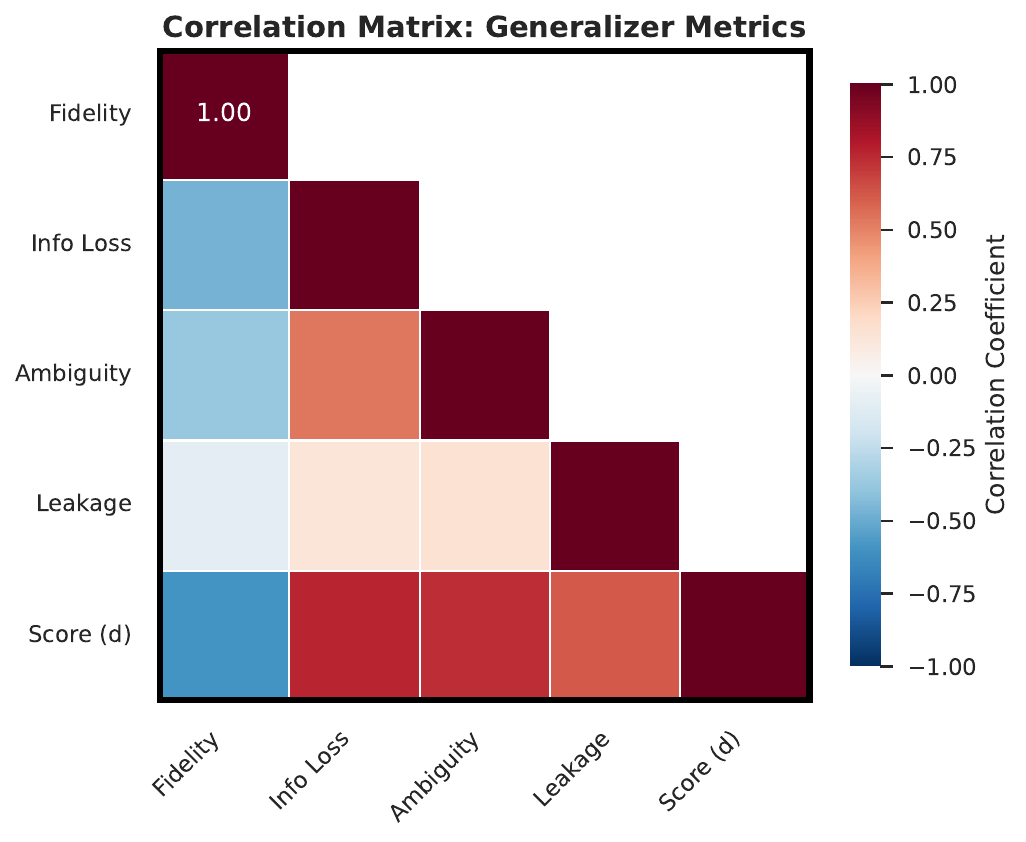}
    \caption{Generalizer quality metric correlations. The deficit score ($d$) correlates strongly with Information Loss and Ambiguity.}
    \label{fig:gen_corr_heatmap}
\end{figure}

Figure~\ref{fig:gen_corr_heatmap} presents the pairwise
correlations among Generalizer quality dimensions, confirming that the deficit score $d$ is primarily driven
by Information Loss and Ambiguity.

\section{Generalizer Metrics}
\label{app:gen_metrics}
Table~\ref{tab:gen_metrics} reports the full Generalizer quality metrics across all nine model configurations.

\begin{table*}[h!]
\centering
\setlength{\tabcolsep}{5pt}
\renewcommand{\arraystretch}{1.06}
\resizebox{0.8\textwidth}{!}{%
\begin{tabular}{@{}ll rrrrrr@{}}
\toprule
\multicolumn{2}{c}{\textbf{Model Configuration}} &
\textbf{Fidelity $\uparrow$} &
\textbf{Info Loss $\downarrow$} &
\textbf{Ambiguity $\downarrow$} &
\textbf{Leakage $\downarrow$} &
\textbf{$d$ $\downarrow$} \\
\cmidrule(r){1-2} \cmidrule(l){3-7}
\textbf{External Model} & \textbf{Internal Model} & & & & & \\
\midrule
\multirow{3}{*}{GPT-OSS-120B}
  & GPT-OSS-120B & 8.80 ± 0.77 & 3.72 ± 1.24 & 3.08 ± 1.11 & 2.16 ± 1.55 & 2.54 ± 0.76 \\
  & Mistral-24B & 8.39 ± 0.85 & 5.24 ± 1.28 & 5.41 ± 0.85 & 1.96 ± 1.45 & 3.56 ± 0.70 \\
  & Qwen-235B & 8.86 ± 0.49 & 4.24 ± 1.92 & 3.03 ± 1.40 & 1.68 ± 1.31 & 2.52 ± 0.94 \\
\midrule
\multirow{3}{*}{Mistral-24B}
  & GPT-OSS-120B & 8.82 ± 0.66 & 3.70 ± 1.16 & 3.05 ± 1.03 & 2.19 ± 1.57 & 2.53 ± 0.70 \\
  & Mistral-24B & 8.41 ± 0.82 & 5.16 ± 1.30 & 5.39 ± 0.80 & 1.89 ± 1.39 & 3.51 ± 0.66 \\
  & Qwen-235B & 8.88 ± 0.47 & 4.19 ± 1.91 & 3.07 ± 1.42 & 1.65 ± 1.23 & 2.51 ± 0.93 \\
\midrule
\multirow{3}{*}{Qwen-235B}
  & GPT-OSS-120B & 8.79 ± 0.80 & 3.71 ± 1.24 & 3.06 ± 1.12 & 2.17 ± 1.56 & 2.54 ± 0.76 \\
  & Mistral-24B & 8.41 ± 0.85 & 5.18 ± 1.34 & 5.39 ± 0.87 & 1.97 ± 1.54 & 3.53 ± 0.74 \\
  & Qwen-235B & 8.87 ± 0.46 & 4.23 ± 1.91 & 3.02 ± 1.32 & 1.63 ± 1.22 & 2.50 ± 0.89 \\
\bottomrule
\end{tabular}%
}
\caption{Generalizer metrics by model configuration. The deficit score ($d\downarrow$) quantifies quality degradation as the average penalty across four dimensions. The large-scale models demonstrate superior performance.}
\label{tab:gen_metrics}
\end{table*}

\section{Performance Across Paper Tiers}
\label{app:paper_tiers}
\begin{figure*}[h!]
    \centering
    \includegraphics[width=0.9\textwidth]{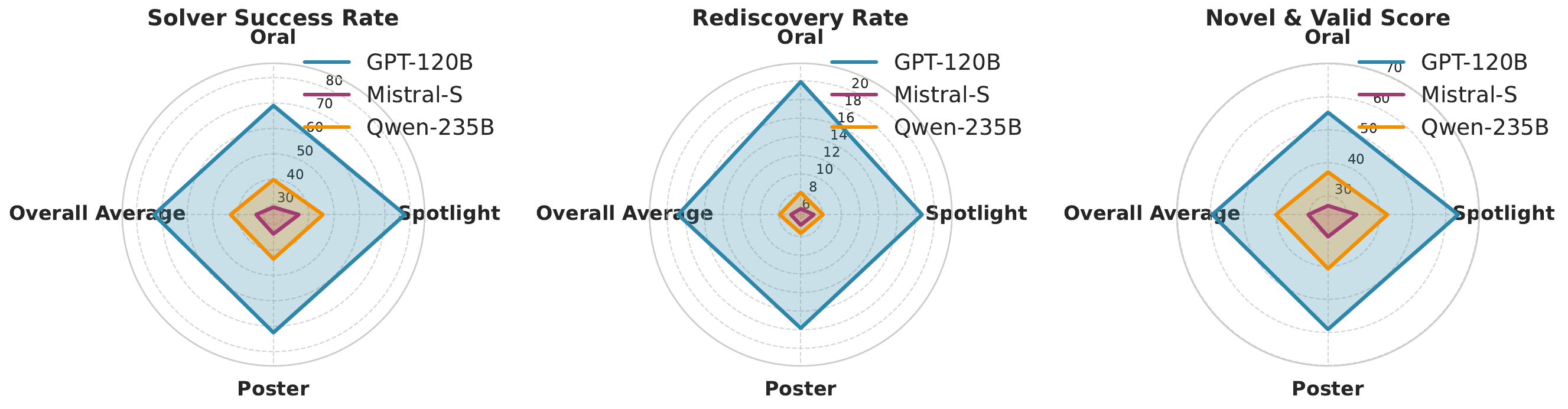}
    \caption{Performance comparison of internal models across ICLR paper tiers (Oral, Spotlight, Poster), averaging across \texttt{GPT-OSS-120B} as the external model ($\tau=5$). The performance hierarchy is stable across tiers.}
    \label{fig:performance_by_tier}
\end{figure*}

Figure~\ref{fig:performance_by_tier} compares Solver
performance stratified by ICLR acceptance tier (Oral,
Spotlight, Poster), demonstrating that the model hierarchy
is stable across paper quality levels.

\section{Visual Analysis of Solution Clusters}
\label{app:visual_clusters}
\begin{figure*}[h!]
    \centering
    \begin{subfigure}[b]{0.60\textwidth}
        \centering
        \includegraphics[width=1.05\textwidth]{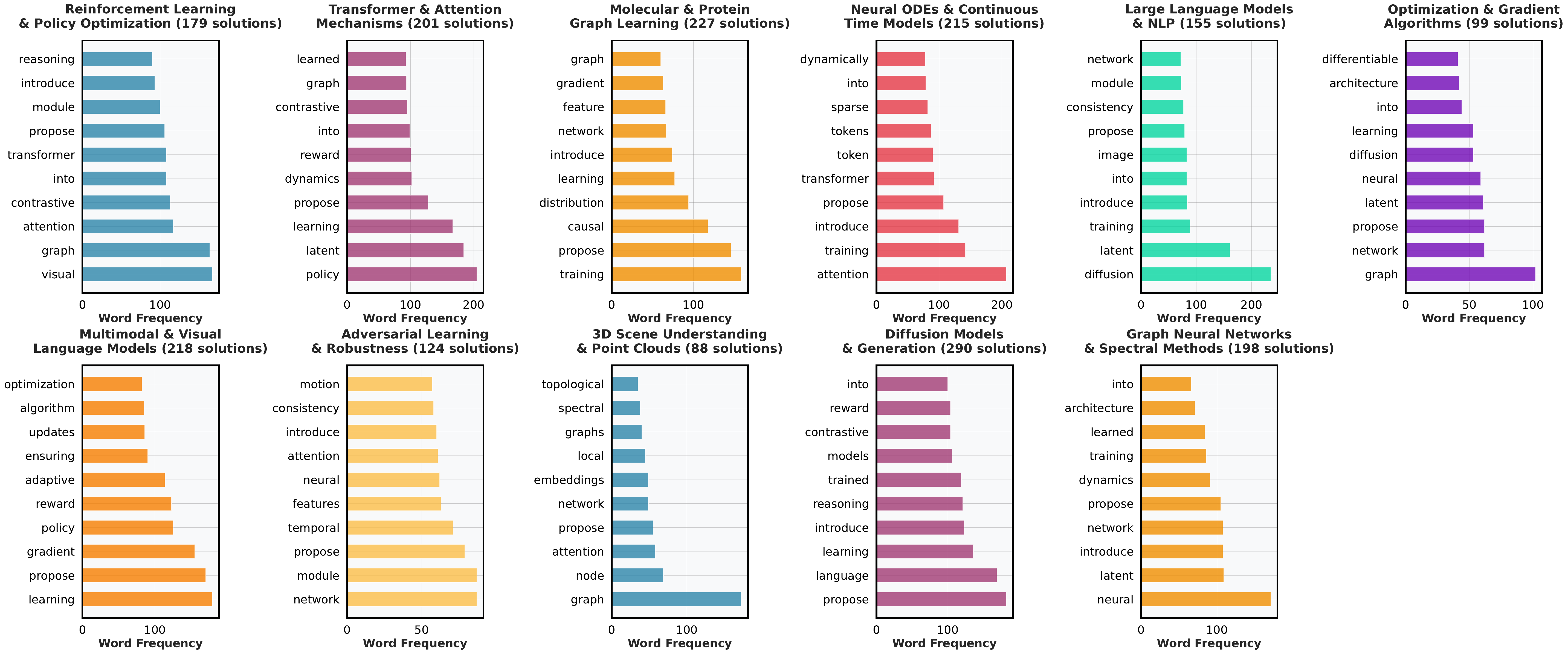}
        \caption{Keyword frequency for prominent clusters.}
        \label{fig:keywords_app}
    \end{subfigure}
    \hfill
    \begin{subfigure}[b]{0.35\textwidth}
        \centering
        \vspace{-0.5em}
        \includegraphics[width=0.92\textwidth]{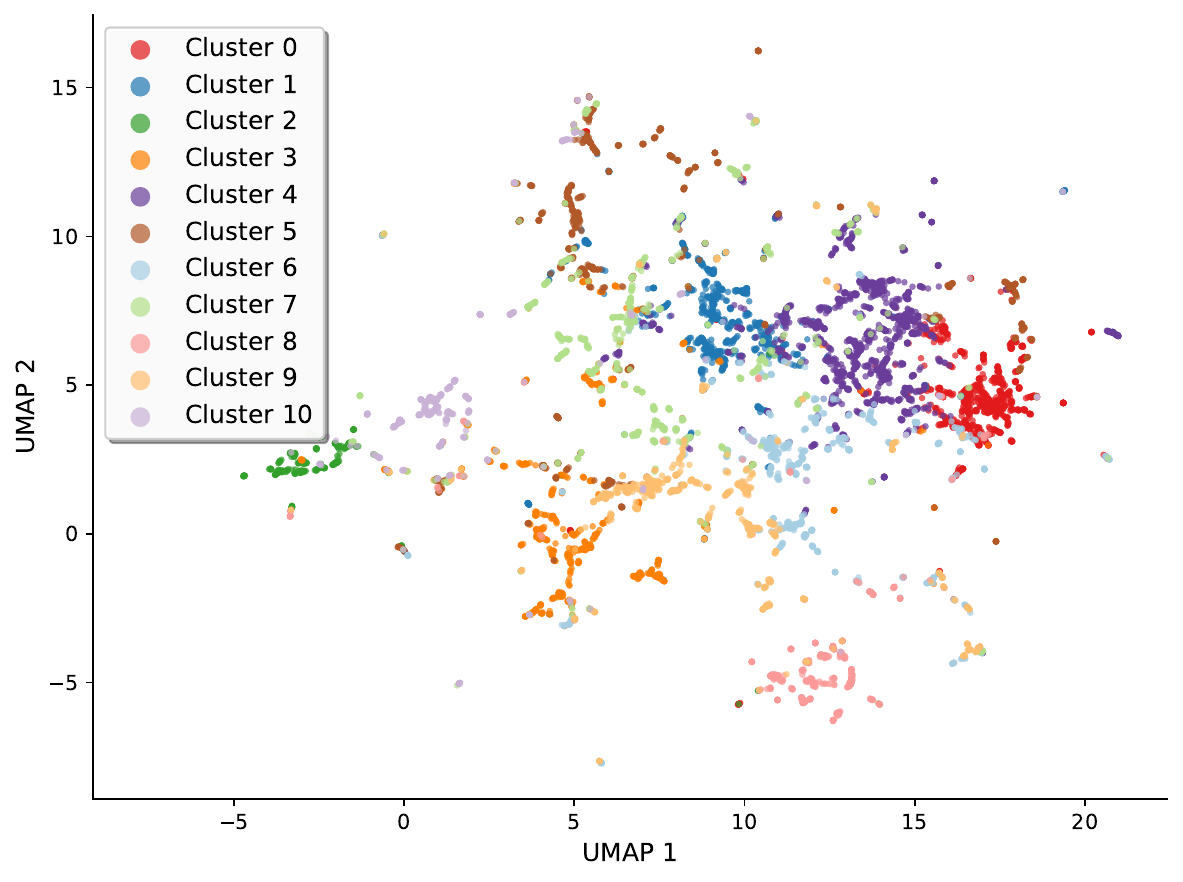}
        \caption{UMAP visualization of solution embeddings.}
        \label{fig:umap_app}
    \end{subfigure}
    \caption{Visual analysis of the 11 identified research clusters. (a) Top keywords provide a thematic summary. (b) The UMAP projection shows distinct grouping by theme.}
    \label{fig:visual_analysis}
\end{figure*}

Figure~\ref{fig:visual_analysis} visualizes the 11 solution
archetypes identified via KMeans++ clustering on Qwen-8B
embeddings.

\section{Statistical Significance Testing}
\label{app:sig_testing}
To rigorously compare the performance between our primary models, \texttt{GPT-OSS-120B} and \texttt{Mistral-24B}, we performed two-sample t-tests and non-parametric Mann-Whitney U tests. We applied a Bonferroni correction to account for multiple comparisons, resulting in a corrected significance threshold of $\alpha \approx 0.0036$.
The results are summarized in Table~\ref{tab:sig_results_main_no_tp}. While we find statistically significant differences in readability metrics, for the core Generalizer metrics, we find no significant differences. We conclude that while \texttt{GPT-OSS-120B} may have a slight, statistically detectable edge, the two models perform at a broadly comparable level in our framework. Thus, motivating our choice of considering problem statements from multiple sources.

\begin{table}[h!]
\centering

\sisetup{
    round-mode=places,
    round-precision=3
}
\setlength{\tabcolsep}{0pt} 

\begin{tabular*}{\textwidth}{@{\extracolsep{\fill}} ll S[table-format=1.2] S[table-format=1.2] S[table-format=1.3] S[table-format=1.3] S[table-format=-1.3] @{}}
\toprule
\textbf{Category} & \textbf{Metric} & {Mean (M)} & {Mean (G)} & {$p$ (t-test)} & {$p$ (MW-U)} & {Cohen's d} \\
\midrule
\multirow{4}{*}{Generalizer} 
 & Semantic Fidelity     & 8.61 & 8.60 & 0.468 & 0.898 & 0.021 \\
 & Information Loss      & 4.43 & 4.48 & 0.254 & 0.288 & -0.033 \\
 & Ambiguity             & 4.22 & 4.25 & 0.601 & 0.659 & -0.015 \\
 & Solution Leakage      & 2.04 & 2.06 & 0.602 & 0.207 & -0.015 \\
\midrule
Readability
 & Flesch-Kincaid Grade* & 22.27 & 21.97 & 0.002 & 0.001 & 0.070 \\
\bottomrule
\end{tabular*}
\caption{Significance testing for key metrics comparing Mistral-24B (M) and GPT-OSS-120B (G). An asterisk (*) denotes statistical significance at the Bonferroni-corrected level ($\alpha \approx 0.0036$). Despite some statistical significance, all effect sizes are negligible.}
\label{tab:sig_results_main_no_tp}
\end{table}

\section{Robustness: Qwen-235B as Evaluator}
\label{app:qwen_eval}

To verify that our findings are not an artifact of the
primary evaluator, we replicate the full Solver evaluation
using Qwen-235B as the LLM judge.
Table~\ref{tab:solver_results_qwen_eval} reports results at
the strict threshold ($\tau\!=\!5$). The performance hierarchy is preserved in its
entirety: GPT-OSS-120B as $\mathcal{M}_i$ consistently
achieves the highest scores across all configurations and
problem sources, regardless of evaluator identity.
\begin{table*}[ht!]
\centering
\scriptsize
\setlength{\tabcolsep}{4pt}
\renewcommand{\arraystretch}{0.99}
\begin{tabular}{@{}ll cc@{}}
\toprule
\multicolumn{2}{c}{\textbf{Model Configuration}} & \textbf{Rediscovery} & \textbf{SR Solver} \\
\cmidrule(r){1-2}\cmidrule(l){3-4}
\textbf{External ($\mathcal{M}_e$)} & \textbf{Internal ($\mathcal{M}_i$)} & & \\
\midrule
\multicolumn{4}{l}{\textbf{Problem Source: GPT-OSS-120B}} \\
\midrule
\multirow{3}{*}{GPT-OSS-120B} & \textbf{GPT-OSS-120B} & \textbf{17.88} & \textbf{84.12} \\
 & Mistral-24B & 6.41 & 50.48 \\
& Qwen-235B & 7.80 & 66.98 \\
\midrule
\multirow{3}{*}{Mistral-24B} & \textbf{GPT-OSS-120B} & \textbf{17.02} & \textbf{81.90} \\
 & Mistral-24B & 6.78 & 49.13 \\
 & Qwen-235B & 7.73 & 62.81 \\
 \midrule
\multirow{3}{*}{Qwen-235B} & \textbf{GPT-OSS-120B} & \textbf{17.26} & \textbf{83.43} \\
 & Mistral-24B & 6.72 & 51.37 \\
 & Qwen-235B & 7.73 & 65.13 \\
\midrule
\multicolumn{4}{l}{\textbf{Problem Source: Mistral-24B}}\\
\midrule
\multirow{3}{*}{GPT-OSS-120B} & \textbf{GPT-OSS-120B} & \textbf{15.88} & \textbf{86.97} \\
& Mistral-24B & 7.10 & 55.98 \\
 & Qwen-235B & 8.63 & 70.30 \\
 \midrule
\multirow{3}{*}{Mistral-24B} & \textbf{GPT-OSS-120B} & \textbf{15.27} & \textbf{86.51} \\
& Mistral-24B & 7.93 & 56.81 \\
& Qwen-235B & 7.82 & 67.08 \\
\midrule
\multirow{3}{*}{Qwen-235B} & \textbf{GPT-OSS-120B} & \textbf{15.24} & \textbf{87.10} \\
 & Mistral-24B & 7.74 & 58.65 \\
 & Qwen-235B & 8.08 & 68.39 \\
\midrule
\multicolumn{4}{l}{\textbf{Problem Source: Qwen-235B}}\\
\midrule
\multirow{3}{*}{GPT-OSS-120B} & \textbf{GPT-OSS-120B} & \textbf{15.65} & \textbf{87.61} \\
& Mistral-24B & 5.86 & 56.55 \\
& Qwen-235B & 7.11 & 70.41 \\
\midrule
\multirow{3}{*}{Mistral-24B} & \textbf{GPT-OSS-120B} & \textbf{14.85} & \textbf{85.99} \\
& Mistral-24B & 5.00 & 54.71 \\
& Qwen-235B & 6.85 & 67.97 \\
\midrule
\multirow{3}{*}{Qwen-235B} & \textbf{GPT-OSS-120B} & \textbf{14.58} & \textbf{87.08} \\
 & Mistral-24B & 6.90 & 58.08 \\
 & Qwen-235B & 6.99 & 69.95 \\
\bottomrule
\end{tabular}
\caption{LLM-as-a-judge success rates (\%) using \textbf{GPT-OSS-120B as the evaluator}. This table validates that the findings from the primary evaluator (Table~\ref{tab:solver_results}) are robust. Bold indicates the best-performing internal model for each external critic.}
\label{tab:solver_results_qwen_eval}
\end{table*}

\section{Full Solver Results Across Problem Sources}
\label{app:full_solver}

Table~\ref{tab:solver_results} reports the complete Solver
performance at $\tau\!=\!5$ across all three problem sources.
Table~\ref{tab:solver_full} reports
only the GPT-OSS-120B problem source; the pattern is
consistent across all sources. In every configuration,
GPT-OSS-120B as $\mathcal{M}_i$ dominates, and
$\mathcal{M}_e$ variation accounts for at most 4 percentage points across
all metrics.

\begin{table*}[h!]
\centering
\scriptsize
\setlength{\tabcolsep}{5pt}
\renewcommand{\arraystretch}{1.0}
\begin{tabular}{@{}ll cc @{}}
\toprule
\multicolumn{2}{c}{\textbf{Model Configuration}}
  & \textbf{Rediscovery}
  & \textbf{SR Solver} \\
\cmidrule(r){1-2}\cmidrule(l){3-4}
\textbf{$\mathcal{M}_e$}
  & \textbf{$\mathcal{M}_i$} & & \\
\midrule
\multicolumn{4}{l}{\textbf{Problem Source: GPT-OSS-120B}} \\
\midrule
\multirow{3}{*}{GPT-OSS-120B}
  & \textbf{GPT-OSS-120B} & \textbf{19.11} & \textbf{74.05} \\
  & Qwen-235B    & 7.74  & 43.82 \\
  & Mistral-24B  & 6.43  & 34.60 \\
\addlinespace[1pt]
\multirow{3}{*}{Mistral-24B}
  & \textbf{GPT-OSS-120B} & \textbf{17.79} & \textbf{71.91} \\
  & Qwen-235B    & 7.17  & 39.21 \\
  & Mistral-24B  & 6.75  & 31.22 \\
\addlinespace[1pt]
\multirow{3}{*}{Qwen-235B}
  & \textbf{GPT-OSS-120B} & \textbf{17.38} & \textbf{70.35} \\
  & Qwen-235B    & 7.66  & 42.34 \\
  & Mistral-24B  & 7.08  & 34.84 \\
\midrule
\multicolumn{4}{l}{\textbf{Problem Source: Mistral-24B}} \\
\midrule
\multirow{3}{*}{GPT-OSS-120B}
  & \textbf{GPT-OSS-120B} & \textbf{16.56} & \textbf{78.01} \\
  & Qwen-235B    & 8.40  & 50.16 \\
  & Mistral-24B  & 7.58  & 42.42 \\
\addlinespace[1pt]
\multirow{3}{*}{Mistral-24B}
  & \textbf{GPT-OSS-120B} & \textbf{16.97} & \textbf{79.41} \\
  & Qwen-235B    & 7.91  & 49.09 \\
  & Mistral-24B  & 8.15  & 40.86 \\
\addlinespace[1pt]
\multirow{3}{*}{Qwen-235B}
  & \textbf{GPT-OSS-120B} & \textbf{15.57} & \textbf{76.28} \\
  & Qwen-235B    & 9.14  & 49.84 \\
  & Mistral-24B  & 8.73  & 43.57 \\
\midrule
\multicolumn{4}{l}{\textbf{Problem Source: Qwen-235B}} \\
\midrule
\multirow{3}{*}{GPT-OSS-120B}
  & \textbf{GPT-OSS-120B} & \textbf{14.91} & \textbf{77.51} \\
  & Qwen-235B    & 6.43  & 50.25 \\
  & Mistral-24B  & 5.11  & 41.85 \\
\addlinespace[1pt]
\multirow{3}{*}{Mistral-24B}
  & \textbf{GPT-OSS-120B} & \textbf{14.74} & \textbf{77.10} \\
  & Qwen-235B    & 6.84  & 49.75 \\
  & Mistral-24B  & 4.37  & 38.88 \\
\addlinespace[1pt]
\multirow{3}{*}{Qwen-235B}
  & \textbf{GPT-OSS-120B} & \textbf{14.09} & \textbf{77.18} \\
  & Qwen-235B    & 6.75  & 50.91 \\
  & Mistral-24B  & 6.34  & 41.02 \\
\bottomrule
\end{tabular}
\caption{Full Solver performance (\%)
all problem sources. Bold = best $\mathcal{M}_i$ per
$\mathcal{M}_e$ block. Note that the problem source denotes the LLM backbone used for problem generation}
\label{tab:solver_full}
\end{table*}

\section{Case Studies}
\label{app:case_studies}

We present three representative examples that illustrate
the spectrum of LLM problem-solving observed in our
evaluation. For each case, we provide the problem
statement, the published human solution, and both LLM
solutions.

\subsection{Successful Rediscovery: Hyperbolic Associative
Memory (Problem~43)}

\paragraph{Problem Statement.}
How can associative memory models be designed to accurately
capture and retrieve hierarchical relationships in data,
given that Euclidean representations distort deep tree-like
structures?

\paragraph{Published Solution.}
The original paper extends modern Hopfield retrieval to
hyperbolic space, mapping queries and memories via
exponential maps, defining the energy function using the
Minkowski inner product, and performing retrieval via
Riemannian gradient descent on the Poincar\'e ball.

\paragraph{GPT-OSS-120B Solution.}
Proposes Hyperbolic Associative Memory (H-AM) on the
Lorentz manifold, replacing dot-product similarity with
Lorentzian inner products and employing Riemannian
optimization with retraction onto the manifold.
Additionally proposes M\"obius addition for parameter
updates.

\paragraph{Mistral-24B Solution.}
Proposes a Hyperbolic Continuous Hopfield Network on the
Poincar\'e ball with M\"obius linear layers, a Lorentzian
inner-product energy function, and Riemannian gradient
descent for retrieval.

\paragraph{Analysis.}
Both models independently recover the core non-trivial
insight of extending Hopfield retrieval to hyperbolic
geometry without access to the published solution. The
key technical components (Lorentzian energy, Riemannian
retrieval dynamics, exponential and M\"obius maps) align
across all three solutions. This constitutes a clean
example of strict Rediscovery ($\tau\!=\!5$), demonstrating
that LLMs can reconstruct non-obvious geometric
constructions from a problem statement alone.

\subsection{Valid but Conventional: GNN Calibration with
OOD Robustness (Problem~20)}

\paragraph{Problem Statement.}
How can the calibration and out-of-distribution detection
capabilities of Graph Neural Networks be enhanced while
preserving predictive accuracy and architectural
flexibility on resource-constrained devices?

\paragraph{Published Solution.}
The original paper adopts a cross-domain approach: spiking
neural networks combined with predictive coding for graph
learning (SIGHT). This replaces backpropagation with local,
spike-driven error correction, a biologically grounded
mechanism that naturally produces calibrated uncertainty and
is inherently event-driven, making it suitable for
power-constrained hardware.

\paragraph{GPT-OSS-120B Solution.}
Proposes Bayesian-Calibrated Message Passing with
variational mean-variance propagation and a
Temperature-Scaled Dirichlet output layer, combined with
Outlier-Exposure via Graph Mixup for OOD robustness.

\paragraph{Mistral-24B Solution.}
Proposes Dirichlet Prior Networks with edge-wise
Monte-Carlo dropout and low-rank variational weight
distributions, augmented by on-the-fly synthetic graph
perturbation for OOD training.

\paragraph{Analysis.}
Both solutions are technically sound and represent
established approaches to uncertainty quantification in
graph neural networks. However, neither model makes the
creative cross-domain transfer to neuroscience-inspired
computation that characterizes the published solution. This
case illustrates how LLMs default to well-explored
methodological paths and producing valid solutions while missing the type of
interdisciplinary conceptual leap that can yield genuinely
surprising approaches.

\subsection{Missed Cross-Domain Transfer: Accelerating AI
Agent Execution (Problem~48)}

\paragraph{Problem Statement.}
How can the execution latency of AI agents be significantly
reduced while guaranteeing that their behavior remains
exactly correct, given that each action requires a
sequential API call?

\paragraph{Published Solution.}
The original paper draws an analogy from CPU speculative
execution and speculative decoding in LLM inference: a
fast, lightweight model predicts likely next actions, which
are executed speculatively in parallel. The results are
then verified against the true policy, with incorrect
predictions discarded, thus yielding lossless acceleration.

\paragraph{GPT-OSS-120B Solution.}
Proposes deterministic result memoization combined with
multi-step batch compilation: caching intermediate outputs
and unrolling sequential calls into batched execution.

\paragraph{Mistral-24B Solution.}
Proposes a Deterministic Local Execution Engine that
compiles agent policies into local ONNX runtimes, reducing
API call overhead through ahead-of-time compilation.

\paragraph{Analysis.}
Both solutions propose implementation-level optimizations
that address the latency symptom but do not recover the
architectural insight of speculative execution. The concept
of predict-and-verify parallelism is well established in
computer architecture, but transplanting it to a novel
context that represents exactly the
type of cross-domain analogical reasoning at which LLMs consistently underperform. This case exemplifies the
central limitation identified throughout our evaluation:
LLMs assemble known techniques effectively within a domain
but rarely transfer concepts across disciplinary
boundaries.

\section{ELO Rankings}
\label{app:elo}

As a complementary validation, we compute ELO ratings from
two independent sources of pairwise preferences.
The authors conducted a head-to-head tournament
on a held-out subset, evaluating randomized, anonymized
solution pairs. Table~\ref{tab:elo_author} reports the
resulting ratings. The GPT-OSS-120B self-play configuration
ranks highest among LLM agents (ELO~1{,}119), second only
to ground-truth human abstracts (ELO~1{,}187). The
consistency between human-derived and author-verified
rankings provides additional evidence that the performance
hierarchy is robust.

\begin{table}[h]
\centering
\small
\setlength{\tabcolsep}{4pt}
\renewcommand{\arraystretch}{1.1}
\begin{tabular}{@{}lrrrr@{}}
\toprule
\textbf{Agent}
  & \textbf{ELO}
  & \textbf{W}
  & \textbf{L}
  & \textbf{Win\%} \\
\midrule
Human Abstracts
  & 1{,}187 & 17 & 1 & 94.4 \\
\textbf{GPT-OSS-120B + GPT-OSS-120B}
  & \textbf{1{,}119} & \textbf{14} & \textbf{5} & \textbf{63.6} \\
Mistral-24B + GPT-OSS-120B
  & 939 & 5 & 8 & 25.0 \\
Mistral-24B + Mistral-24B
  & 927 & 5 & 12 & 25.0 \\
GPT-OSS-120B + Mistral-24B
  & 828 & 1 & 16 & 5.0 \\
\bottomrule
\end{tabular}
\caption{ELO ratings from pairwise
preferences on a held-out subset (as verified by authors).}
\label{tab:elo_author}
\end{table}

\paragraph{Convergence Between Rankings.}
Both ELO rankings converge on the same conclusions:
(i)~LLM agents are competitive with human abstracts,
(ii)~GPT-OSS-120B as $\mathcal{M}_i$ is the dominant
factor, and (iii)~the author-verified ranking
(GPT~$1{,}119$ $>$ Mistral~$939$) exhibits a wider gap
than the human-derived ranking
(GPT~$1{,}046$ $>$ Mistral~$1{,}020$).

This appendix provides a selection of head-to-head comparisons from our human-verified ELO tournament. For each example, we present the problem statement, the responses from the two competing agents, and the final verdict from the human evaluator.

\vspace{1em} 

\paragraph{Example 1: Video Inverse Problems (ID: TRWxFUzK9K)}
\begin{itemize}
    \item \textbf{Problem Statement:} How can we formulate and solve video inverse problems that involve spatio-temporal degradations using only pre-trained image diffusion models, while ensuring consistent reconstruction across the temporal dimension without training dedicated video diffusion models?
    \item \textbf{Agent A Response (\texttt{Mistral-24B} Self-Play):} We propose a method called \textbf{Temporal Consistency-aware Recurrent Image Diffusion (TC-RID)} that adapts pre-trained image diffusion models for video inverse problems. TC-RID introduces a recurrent neural network (RNN) that processes each video frame sequentially, using the hidden state to maintain temporal information...
    \item \textbf{Agent B Response (\texttt{GPT-OSS-120B} Self-Play):} We formulate video inverse problems as a sequence of frame-wise diffusion processes in which a frozen pre-trained image diffusion model acts as the denoiser, but each denoising step receives an additional cross-frame guidance tensor obtained by warping the latent of the previous frame with a fast optical-flow estimator...
    \item \textbf{Winner: B}. Agent B's solution was judged to be more technically specific and innovative by proposing the use of optical flow and warped latents as a temporal prior.
\end{itemize}

\hrule\vspace{1em}

\paragraph{Example 2: Dynamic Mixture of Experts (ID: T26f9z2rEe)}
\begin{itemize}
    \item \textbf{Problem Statement:} How can a Mixture of Experts architecture automatically determine the appropriate number of experts to activate for each input token and dynamically adjust the total number of experts during training, thereby eliminating the need for extensive hyper-parameter search?
    \item \textbf{Agent A Response (Human Abstract):} The Sparse Mixture of Experts (SMoE) has been widely employed... However, the performance of SMoE heavily depends on the choice of hyper-parameters... As a remedy, we introduce the Dynamic Mixture of Experts (DynMoE) technique. DynMoE incorporates (1) a novel gating method that enables each token to automatically determine the number of experts to activate...
    \item \textbf{Agent B Response (\texttt{GPT-OSS-120B} Self-Play):} We introduce \textbf{Dynamic Adaptive Gating with an Elastic Expert Pool (DAGE-E)}: a global gating network outputs a full softmax over a large superset of experts, while a learned token-wise budget scalar determines a differentiable top-k cutoff... Concurrently, an Elastic Expert Pool manager monitors per-expert utilization statistics...
    \item \textbf{Winner: A (Human)}. While the LLM's proposal (DAGE-E) was highly detailed and technically plausible, the human abstract was judged to be a slightly more direct and elegant solution to the core problem.
\end{itemize}

\hrule\vspace{1em}

\paragraph{Example 3: Configurable LLM Safety (ID: ERce2rgMQC)}
\begin{itemize}
    \item \textbf{Problem Statement:} How can large language models be equipped with a mechanism that allows their safety behavior to be dynamically configured at inference time to satisfy diverse cultural, regional, and individual safety requirements, without requiring model retraining?
    \item \textbf{Agent A Response (\texttt{GPT-OSS-120B} Self-Play):} We introduce a \textbf{Dynamic Safety Router (DSR)} that sits between the transformer layers of a frozen LLM and its output head... the DSR computes per-token gating vectors that multiplicatively modulate the hidden state...
    \item \textbf{Agent B Response (\texttt{Mistral-24B} $\to$ \texttt{GPT-OSS-120B}):} We introduce a \textbf{Safety-Conditioned HyperAdapter (SCH)} architecture that sits between a frozen LLM and its output: a lightweight hypernetwork receives a structured safety profile and instantly emits low-rank additive weight matrices for each transformer layer...
    \item \textbf{Winner: B}. The HyperAdapter approach was judged to be a more modern and flexible architecture for inference-time configuration compared to the gating mechanism.
\end{itemize}

\hrule\vspace{1em}

\paragraph{Example 4: Accelerating Diffusion Models (ID: OlzB6LnXcS)}
\begin{itemize}
    \item \textbf{Problem Statement:} How can we substantially accelerate the sampling process of diffusion and flow-matching generative models while preserving or improving sample quality, using a single network trained in a single phase?
    \item \textbf{Agent A Response (\texttt{Mistral-24B} Self-Play):} We propose a novel architecture, \textbf{Sparse Adaptive Flow Network (SAFNet)}, which integrates adaptive sparse convolutions with dynamic skip connections...
    \item \textbf{Agent B Response (\texttt{GPT-OSS-120B} Self-Play):} We introduce \textbf{Dynamic-Step Diffusion (DSD)}: a continuous-time diffusion model that receives... a learned embedding of the desired step size... During training we... minimize a combined loss: (i) the standard score/flow matching loss... and (ii) a consistency loss...
    \item \textbf{Winner: B}. The DSD proposal with its consistency loss was deemed a more theoretically grounded and powerful approach for handling variable step sizes compared to SAFNet.
\end{itemize}

\hrule\vspace{1em}

\paragraph{Example 5: Stochastic Linear Bandits (ID: Oeb0I3JcVc)}
\begin{itemize}
    \item \textbf{Problem Statement:} Is it possible to derive instance-specific, data-driven regret guarantees for stochastic linear bandit algorithms... and to adjust these algorithms so that they preserve their empirical effectiveness while achieving the minimax optimal regret?
    \item \textbf{Agent A Response (\texttt{Mistral-24B} Self-Play):} We propose an adaptive, instance-specific regret-guaranteed stochastic linear bandit algorithm \textbf{ISRG-SLB} that dynamically adjusts its exploration-exploitation trade-off... ISRG-SLB employs a Bayesian posterior sampling mechanism...
    \item \textbf{Agent B Response (\texttt{Mistral-24B} $\to$ \texttt{GPT-OSS-120B}):} We introduce \textbf{Hardness-Adaptive Regret Guard (HARG)}, a meta-algorithm that runs a base linear bandit method... but continuously computes an instance-specific hardness score... When the score exceeds a threshold, HARG switches to a robust OFUL-style exploration policy...
    \item \textbf{Winner: C (Tie/Cannot Decide)}. The evaluator judged both proposals to be highly sophisticated and plausible meta-algorithms that addressed the core problem in different but equally valid ways, making a clear winner difficult to determine.
\end{itemize}

\hrule\vspace{1em}

\paragraph{Example 6: Root-Cause Analysis (ID: l11DZY5Nxu)}
\begin{itemize}
    \item \textbf{Problem Statement:} How can we reliably identify the true root-cause nodes of an anomaly in a complex system when anomalies are rare and lie outside the training distribution of a structural causal model, making counterfactual estimates unreliable?
    \item \textbf{Agent A Response (\texttt{Mistral-24B} $\to$ \texttt{GPT-OSS-120B}):} We introduce \textbf{Distribution-Aware Counterfactual Projection (DACP)}: (1) train a causal invertible generative model... on abundant normal operation data... (2) given an anomalous observation, compute its in-distribution projection... (3) for each candidate... perform an on-manifold do-intervention...
    \item \textbf{Agent B Response (\texttt{GPT-OSS-120B} $\to$ \texttt{Mistral-24B}):} We propose a novel approach called \textbf{Causal Influence Propagation (CIP)} that operates within the in-distribution regime by iteratively simulating the propagation of influences through the causal graph. Given an observed anomaly, CIP initializes a set of candidate root-cause nodes...
    \item \textbf{Winner: A}. The DACP proposal was judged to be a more rigorous and theoretically grounded approach by explicitly using a generative model to project anomalies back to the in-distribution manifold before intervention.
\end{itemize}

\hrule\vspace{1em}

\paragraph{Example 7: Promptable 3D Segmentation (ID: yXCTDhZDh6)}
\begin{itemize}
    \item \textbf{Problem Statement:} How can a unified, scalable foundation model be developed for promptable 3-D segmentation of point clouds that overcomes heterogeneous data formats and the scarcity of diverse, richly annotated 3-D mask data?
    \item \textbf{Agent A Response (\texttt{Mistral-24B} $\to$ \texttt{GPT-OSS-120B}):} We introduce \textbf{GeoPrompt-3D}, a unified sparse-voxel transformer that first tokenizes any point-cloud format into a geometry-aware token lattice... A multimodal prompt encoder maps textual descriptions, point clicks, or exemplar masks into a set of prompt tokens...
    \item \textbf{Agent B Response (Human Abstract):} The development of 2D foundation models for image segmentation has been significantly advanced by the Segment Anything Model (SAM). However, achieving similar success in 3D models remains a challenge... To this end, we propose a 3D promptable segmentation model Point-SAM... We then distill the rich knowledge from 2D SAM for Point-SAM training...
    \item \textbf{Winner: B (Human)}. The human-proposed solution of distilling knowledge from the existing 2D SAM was judged to be a more pragmatic and powerful approach to overcoming data scarcity than the LLM's proposal of generating synthetic masks with a diffusion model.
\end{itemize}

\section{Extended Qualitative Analysis of Solution Archetypes}
\label{app:cluster_extend}

\paragraph{Semantic Coherence and Readability.} Semantic coherence metrics show that \texttt{GPT-OSS-120B} solutions achieve high cosine similarity ($\approx 0.87$) with the problems they address, indicating strong conceptual alignment. Readability scores reveal that large-scale models produce more technically dense proposals (Flesch-Kincaid $\approx 23$--$26$) than \texttt{Mistral-24B} ($\approx 22$), consistent with their superior reasoning performance.

\paragraph{Solution Archetypes.}
To understand the conceptual landscape of the generated solutions, we performed a qualitative analysis by clustering the entire solution space. We grouped solutions into semantically coherent clusters via KMeans++ on their embeddings and assigned each cluster an interpretable label based on its most characteristic keywords.

This process reveals a rich taxonomy of 11 distinct solution archetypes, summarized in Table~\ref{tab:cluster_archetypes}. The analysis demonstrates that the agents do not merely produce generic outputs but explore a diverse range of modern research paradigms. These archetypes span broad strategies like Reinforcement Learning (C0), foundational model architectures (C1), and highly specialized domains such as Molecular Graph Modeling (C2) and 3D Scene Representation (C8).

\begin{table*}[h!]
\centering
\caption{Taxonomy of identified solution archetypes. Cohesion is the average intra-cluster cosine similarity.}
\label{tab:cluster_archetypes}
\resizebox{\textwidth}{!}{%
\begin{tabular}{@{}c l >{\raggedright\arraybackslash}p{7.5cm} c c@{}}
\toprule
\textbf{ID} & \textbf{Archetype Label} & \textbf{Top Keywords} & \textbf{Cohesion} & \textbf{Size} \\
\midrule
C0  & Reinforcement Learning      & policy, learning, reward, agent, action                  & 0.48 & 1037 \\
C1  & Transformer Architectures   & attention, model, token, rank, layer                     & 0.46 & 1106 \\
C2  & Molecular Graph Modeling    & protein, graph, molecular, diffusion, equivariant        & 0.48 & 398  \\
C3  & Neural/Latent Dynamics      & neural, latent, time, network, dynamics                  & 0.41 & 899  \\
C4  & LLM Reasoning               & model, llm, language, reasoning, human                   & 0.46 & 1647 \\
C5  & Adaptive Gradient Methods   & gradient, adaptive, algorithm, convergence, learning     & 0.38 & 936  \\
C6  & Multimodal Learning         & visual, modal, language, transformer, video              & 0.46 & 911  \\
C7  & Data-Centric AI             & data, model, training, loss, class                       & 0.41 & 974  \\
C8  & 3D Scene Representation     & 3d, scene, point, view, motion                           & 0.48 & 548  \\
C9  & Diffusion Models            & diffusion, latent, image, model, diffusion model         & 0.51 & 775  \\
C10 & Graph Neural Networks       & graph, node, gnn, edge, graphs                           & 0.47 & 424  \\
\bottomrule
\end{tabular}%
}
\end{table*}

\begin{clusterbox}
{Cluster 1}{LightRoyalBlue}{Transformer Architecture \& Attention (N=1106)}
    \textbf{Keywords:} attention, model, token \\
    \textbf{Representative Solution:} We introduce the \textbf{Meta-Optimized Sparse Mixture-of-Experts Transformer (MOS-MoE)}, in which a lightweight hypernetwork predicts, for each token and layer, a sparse subset (e.g., 2-4) of expert feed-forward networks.
\end{clusterbox}

\begin{clusterbox}{Cluster 4}{LightPurple}{Large Language Models \& NLP (N=1647)}
    \textbf{Keywords:} model, llm, language \\
    \textbf{Representative Solution:} We introduce a hierarchical self-reward framework where a population of heterogeneous critic LLMs (parameter-perturbed copies) generate pairwise preference judgments for candidate completions, and a main LLM is fine-tuned on this preference data.
\end{clusterbox}

The conceptual cohesion of each archetype, measured by intra-cluster cosine similarity, offers further insight. For example, Diffusion Models (C9) form the most semantically coherent group (0.51 similarity), suggesting a well-defined and consistently applied solution pattern. In contrast, Adaptive Gradient Methods (C5) are the most diverse (0.38 similarity), indicating a broader set of varied approaches within that theme. This analysis confirms that the agents are capable of generating a wide spectrum of scientifically relevant and conceptually distinct solutions, providing a qualitative validation of their creative problem-solving abilities.

\paragraph{Representative Solution Archetypes.}
Below we provide a representative generated solution for each of the 11 identified clusters.

\begin{clusterbox}{Cluster 0}{LightRed}{Reinforcement Learning \& Policy Optimization (N=1037)}
    \textbf{Keywords:} policy, learning, reward \\
    \textbf{Representative Solution:} We introduce \textbf{Meta-Regularized Adaptive Normalization (MRAN)}, a model-free actor-critic algorithm that augments a shared Transformer-based policy-value network with a lightweight context encoder.
\end{clusterbox}

\begin{clusterbox}{Cluster 1}{LightRoyalBlue}{Transformer Architecture \& Attention (N=1106)}
    \textbf{Keywords:} attention, model, token \\
    \textbf{Representative Solution:} We introduce the \textbf{Meta-Optimized Sparse Mixture-of-Experts Transformer (MOS-MoE)}, in which a lightweight hypernetwork predicts, for each token and layer, a sparse subset (e.g., 2-4) of expert feed-forward networks.
\end{clusterbox}

\begin{clusterbox}{Cluster 2}{LightGreen}{Molecular \&  Protein Graph Learning (N=398)}
    \textbf{Keywords:} protein, graph, molecular \\
    \textbf{Representative Solution:} We propose a novel \textbf{Equivariant Variational Conformer Autoencoder (EVC-AE)} for structure-based drug design, which extends equivariant graph neural networks (EGNNs) with a variational autoencoder.
\end{clusterbox}

\begin{clusterbox}{Cluster 3}{LightOrange}{Neural ODEs \& Continuous-Time Models (N=899)}
    \textbf{Keywords:} neural, latent, time \\
    \textbf{Representative Solution:} We propose a Temporal Neural Implicit ODE (TNI-ODE) architecture in which a deep network $f_{\theta}(x, t, \Delta t)$ parameterizes a continuous-time vector field that explicitly receives the elapsed time $\Delta t$ between observations.
\end{clusterbox}

\begin{clusterbox}{Cluster 4}{LightPurple}{Large Language Models \& NLP (N=1647)}
    \textbf{Keywords:} model, llm, language \\
    \textbf{Representative Solution:} We introduce a hierarchical self-reward framework where a population of heterogeneous critic LLMs (parameter-perturbed copies) generate pairwise preference judgments for candidate completions, and a main LLM is fine-tuned on this preference data.
\end{clusterbox}

\begin{clusterbox}{Cluster 5}{LightBrown}{Optimization \& Gradient Algorithms (N=936)}
    \textbf{Keywords:} gradient, adaptive, algorithm \\
    \textbf{Representative Solution:} We propose \textbf{Sparse Adaptive Sketch-Based Gradient Estimation (SASGE)}: at each online round the algorithm draws a set of $k = O(s\log d)$ *sparse* random perturbation vectors whose support follows a structured distribution.
\end{clusterbox}

\begin{clusterbox}{Cluster 6}{LighterSkyBlue}{Multimodal \& Visual Language Models (N=911)}
    \textbf{Keywords:} visual, modal, language \\
    \textbf{Representative Solution:} We introduce a novel architecture called \textbf{Compositional Semantic Transformer (CoST)} that integrates a multi-modal encoder with a graph-based reasoning module. The CoST model encodes visual and linguistic inputs into a shared semantic space.
\end{clusterbox}

\begin{clusterbox}{Cluster 7}{LightLimeGreen}{Adversarial Learning \& Robustness (N=974)}
    \textbf{Keywords:} data, model, training \\
    \textbf{Representative Solution:} We propose a \textbf{Decoupled Adversarial Invariance Learning (DAIL)} framework that decouples the training of invariance and classification objectives using a dual-branch architecture with shared representations but separate adversarial heads.
\end{clusterbox}

\begin{clusterbox}{Cluster 8}{LightPeach}{3D Vision \& Scene Representation (N=548)}
    \textbf{Keywords:} 3d, scene, point \\
    \textbf{Representative Solution:} We introduce \textbf{Dynamic Implicit Neural Volume (DINV)}: given a few calibrated input images, a multi-scale patch transformer extracts per-patch embeddings and aggregates them via a cross-attention module to query a dynamic MLP-based radiance field.
\end{clusterbox}

\begin{clusterbox}{Cluster 9}{LightPink}{Generative Diffusion Models (N=775)}
    \textbf{Keywords:} diffusion, latent, image \\
    \textbf{Representative Solution:} We propose a \textbf{Perceptual-Guided Diffusion (PGD) framework} that augments the standard denoising diffusion process with a trainable Perceptual Guidance Module (PGM). At each denoising step $t$, the cross-attention layers of the U-Net receive conditioning from the PGM.
\end{clusterbox}

\begin{clusterbox}{Cluster 10}{LightThistle}{Graph Neural Networks (GNNs) (N=424)}
    \textbf{Keywords:} graph, node, gnn \\
    \textbf{Representative Solution:} We introduce \textbf{Spectral-Diffusion Graph Transformers (SD-GT)}, which augment the standard self-attention with (i) a set of learnable wavelet-filtered Laplacian eigen-vectors that provide multi-scale spectral information and (ii) a graph diffusion process.
\end{clusterbox}

\section{Agent and Critic Prompts}
\label{app:prompts}
This section contains the exact prompts used for the Generalizer agent, the Solver agent, and their corresponding internal/external critics.

\begin{promptbox}{Prompt: Generalizer Agent ($\mathcal{G}$)}
\label{prompt:generalizer}
\small
\textbf{System Role:}
\begin{quote}
You are an AI researcher with 20 years of experience in the field. Your task is to read a research abstract and identify the core research question tackled by the paper. You must be extremely careful to:
\begin{itemize}
    \item Preserve the fundamental scientific challenge
    \item Avoid hinting at specific solution methods
    \item Maintain precision and clarity
\end{itemize}
\end{quote}
\textbf{User Prompt:}
\begin{quote}
\textbf{Original Research Abstract:}\\
\texttt{\{abstract\}}

\vspace{1em}
\textbf{Your Task:}\\
Write the core research question that captures the core scientific problem described in the abstract.

\vspace{1em}
\textbf{Requirements:}
\begin{itemize}
    \item \textbf{Semantic Fidelity}: Preserve the fundamental scientific challenge exactly.
    \item \textbf{Information Preservation}: Retain all critical details, constraints, and insights.
    \item \textbf{Specificity}: Be precise and unambiguous.
    \item \textbf{Solution Blindness}: Do not hint at or describe the specific solution method.
\end{itemize}

\vspace{1em}
\textbf{Output Format:}
\begin{itemize}
    \item Provide your problem statement in 2-3 clear, concise sentences.
    \item Provide a justification for the identified research question.
    \item Enclose the problem statement inside \texttt{<problem\_statement></problem\_statement>} tags.
    \item Enclose your justification within \texttt{<justification></justification>} tags.
\end{itemize}
\end{quote}
\end{promptbox}

\begin{promptbox}{Prompt: Generalizer Agent ($\mathcal{G}$)}
\small
\textbf{System Role:}
\begin{quote}
You are an expert evaluator specializing in assessing the quality of problem statements from research abstracts. Your role is to critically evaluate whether a problem statement successfully captures the core idea of a research abstract across multiple dimensions.
\end{quote}
\textbf{User Prompt:}
\begin{quote}
\textbf{Original Research Abstract:}\\
\texttt{\{original\_abstract\}}

\vspace{1em}
\textbf{Extracted Problem Statement to Evaluate:}\\
\texttt{\{problem\}}

\vspace{1em}
\textbf{Evaluation Task:}\\
Assess the quality of the problem statement against the original abstract using the following dimensions:
\begin{itemize}
    \item \textbf{Semantic Fidelity (1-10):} How well does the problem statement preserve the core scientific problem? (10 = identical challenge).
    \item \textbf{Information Loss (1-10):} Assess the severity of any missing critical details. (10 = critical info lost, 1 = no loss).
    \item \textbf{Ambiguity (1-10):} Rate the ambiguity of the problem. (10 = highly ambiguous, 1 = perfectly specific).
    \item \textbf{Solution Leakage (1-10):} Does it hint at the solution? (10 = explicitly describes solution, 1 = completely blind).
\end{itemize}
Finally, provide a final judgement on whether the problem statement preserves the core problem.

\vspace{1em}
\textbf{Output Format:}\\
Provide your evaluation in a structured format with separate tags for each assessment (\texttt{<semantic\_fidelity\_assessment>}, etc.) and a \texttt{<final\_judgement>} tag.
\end{quote}
\end{promptbox}

\begin{promptbox}{Prompt: Solver Agent ($\mathcal{S}$)}
\label{prompt:solver}
\small
\textbf{System Role:}
\begin{quote}
You are an expert AI research scientist. Your task is to invent a plausible technical approach that could solve a given scientific problem in machine learning. You must be creative and innovative, proposing novel solutions.
\end{quote}
\textbf{User Prompt:}
\begin{quote}
\textbf{Problem Statement:}\\
\texttt{\{problem\}}

\vspace{1em}
\textbf{Your Task:}\\
Propose a specific and novel technical approach, mechanism, or architecture. Explain your proposed method in 3–5 sentences as if you're writing the core idea in a research paper.

\vspace{1em}
\textbf{Requirements:}
\begin{itemize}
    \item \textbf{Novelty \& Creativity}: Propose a non-obvious, innovative solution.
    \item \textbf{Technical Feasibility}: Ensure your approach is logically sound and implementable.
    \item \textbf{Completeness}: Provide enough detail to understand the core methodology.
\end{itemize}

\vspace{1em}
\textbf{Output Format:}
\begin{itemize}
    \item Provide your proposed technical approach in 3-5 sentences.
    \item Provide a brief justification for the proposed solution.
    \item Enclose your solution inside \texttt{<solution></solution>} tags.
    \item Enclose your justification within \texttt{<justification></justification>} tags.
\end{itemize}
\end{quote}
\end{promptbox}

\begin{promptbox}{Prompt: Solution Critic ( for Solver)}
\small
\textbf{System Role:}
\begin{quote}
You are an expert evaluator specializing in assessing the quality of proposed solutions for complex research problems. Your role is to evaluate the proposed solution solely on how well it solves the problem. You must be thorough and objective in your assessment.
\end{quote}
\textbf{User Prompt:}
\begin{quote}
\textbf{Problem Statement:}\\
\texttt{\{problem\}}

\vspace{1em}
\textbf{Proposed Solution:}\\
\texttt{\{pred\}}

\vspace{1em}
\textbf{Evaluation Task:}\\
Assess the quality of the proposed solution using the following dimensions:
\begin{itemize}
    \item \textbf{Novelty \& Creativity (1-10):} How novel is the approach? (10 = highly creative).
    \item \textbf{Technical Feasibility (1-10):} Is the solution technically sound? (10 = perfectly plausible).
    \item \textbf{Completeness \& Detail (1-10):} How complete is the methodology? (10 = fully specified).
\end{itemize}
Finally, write your final judgement indicating whether the proposed solution ultimately solves the problem.

\vspace{1em}
\textbf{Output Format:}\\
Provide your evaluation in a structured format with separate tags for each assessment (\texttt{<novelty\_assessment>}, etc.) and a \texttt{<final\_judgement>} tag.
\end{quote}
\end{promptbox}

\section{Evaluation Prompts}
\label{app:eval_prompts}
This section contains the prompts used by the LLM-as-a-judge for automated evaluation (Success Rate, Rediscovery) and the monolithic reviewer prompt used in the ResearchPlanGen experiments.

\begin{promptbox}{Prompt: Success Rate Evaluator}
\label{prompt:sr_eval}
\small
\textbf{System Role:}
\begin{quote}
You are an expert reviewer for machine learning research. You will be shown a problem statement and a proposed solution and your task is to evaluate if the proposed solution solved the problem using a structured rubric.
\end{quote}
\textbf{User Prompt:}
\begin{quote}
\textbf{Evaluation Task: Solution Success Rate}

You will be shown a problem statement and a proposed solution. Your task is to assess how well the proposed solution directly addresses and solves the problem.

\textbf{Input:}\\
Problem Statement: \texttt{\{gt\}}\\
Proposed Solution: \texttt{\{pred\}}

\textbf{Task Rubric:}
\begin{itemize}
    \item \textbf{5} (Direct \& Complete Solution): Directly and comprehensively solves the problem. All aspects addressed effectively.
    \item \textbf{4} (Strong Alignment): Strongly aligns with the problem and provides a nearly complete solution. Minor aspects may require minimal elaboration.
    \item \textbf{3} (Moderate Alignment): Addresses significant aspects but leaves some key parts unresolved. Noticeable gaps.
    \item \textbf{2} (Limited Alignment): Addresses only peripheral aspects or offers a very incomplete solution.
    \item \textbf{1} (Minimal Alignment): Minimal relevance. May address a tangentially related issue.
    \item \textbf{0} (No Alignment): Entirely unrelated to the problem statement.
\end{itemize}

\textbf{Output:} JSON with \texttt{relevanceScore} (0--5), \texttt{confidenceLevel} (0--10), and \texttt{summaryStatement}.
\end{quote}
\end{promptbox}

\begin{promptbox}{Prompt: Rediscovery Evaluator}
\label{prompt:rediscovery}
\small
\textbf{System Role:}
\begin{quote}
You are an expert reviewer in machine learning research. You will be shown a research abstract and a proposed solution. Your task is to evaluate the conceptual alignment between the solution in the original abstract and the proposed solution.
\end{quote}
\textbf{User Prompt:}
\begin{quote}
\textbf{Task Name: Solution Rediscovery}

Assess the degree to which the proposed solution rediscovers or conceptually aligns with the core idea, methodology, or key findings in the original abstract. Focus on fundamental conceptual overlap, not exact wording.

\textbf{Input:}\\
Research Abstract: \texttt{\{gt\}}\\
Proposed Solution: \texttt{\{pred\}}

\textbf{Task Rubric:}
\begin{itemize}
    \item \textbf{5} (Perfect Rediscovery): Directly articulates the core idea or primary methodology of the original abstract. Strong, explicit conceptual alignment.
    \item \textbf{4} (Strong Alignment): Captures the main conceptual thrust with minor variations in approach or implementation.
    \item \textbf{3} (Moderate Overlap): Shares key conceptual ideas but with notable divergences or only covers a subset of the core problem.
    \item \textbf{2} (Peripheral Alignment): Addresses a related problem or uses broadly similar techniques but does not align with the core contribution.
    \item \textbf{1} (Minimal Relevance): Shares general domain/terminology but no meaningful conceptual overlap.
    \item \textbf{0} (No Relevance): Entirely unrelated domain or ideas.
\end{itemize}

\textbf{Output:} JSON with \texttt{relevanceScore} (0--5), \texttt{confidenceLevel} (0--10), and \texttt{summaryStatement}.
\end{quote}
\end{promptbox}

\begin{promptbox}{Prompt: Reviewer (ResearchPlanGen)}
\label{prompt:monolithic_reviewer}
\small
\textbf{System Role:}
\begin{quote}
You are an expert evaluator specializing in assessing the quality of proposed solutions for complex research problems. Your role is to evaluate the proposed solution on how well it addresses specific evaluation criteria. You must be thorough, strict, and objective.
\end{quote}
\textbf{User Prompt:}
\begin{quote}
\texttt{\{few\_shot\}} \textit{[empty for self-criteria; $k$ examples with rubrics for few-shot conditions]}

\textbf{Problem Statement:} \texttt{\{problem\}}\\
\textbf{Proposed Solution:} \texttt{\{pred\}}\\
\texttt{\{rubric\}} \textit{[self-generated for generic conditions; ground-truth for rubric-aware oracle]}

\textbf{Evaluation Task:}
Assess whether the proposed solution adequately addresses each evaluation criterion. Consider: Does the plan handle each criterion with sufficient implementation detail? Are there overlooked flaws? Is the approach well-justified?

Then provide overall scores for Criteria Coverage (1--10), Technical Feasibility (1--10), and Completeness \& Detail (1--10).

\textbf{Output Format:}\\
\texttt{<rubric\_assessment>} [For each criterion: SATISFIED or UNSATISFIED with reason] \texttt{</rubric\_assessment>}\\
\texttt{<novelty\_assessment>} Score: X/10 \texttt{</novelty\_assessment>}\\
\texttt{<technical\_assessment>} Score: X/10 \texttt{</technical\_assessment>}\\
\texttt{<completeness\_assessment>} Score: X/10 \texttt{</completeness\_assessment>}\\
\texttt{<final\_judgement>} Justification + Judgement: yes/no. Be STRICT: reject solutions that miss more than 2 criteria or only superficially address them. \texttt{</final\_judgement>}
\end{quote}
\end{promptbox}

\section{ResearchPlanGen Experimental Condition Prompts}
\label{app:rpg_prompts}

This section documents the prompt templates that distinguish the experimental conditions in the ResearchPlanGen evaluation (Section~\ref{sec:rpg}). All conditions use the monolithic reviewer prompt (Prompt~\ref{prompt:monolithic_reviewer}) as the base, with the \texttt{\{rubric\}} and \texttt{\{few\_shot\}} template variables populated differently.

\begin{promptbox}{Self-Criteria Generation (Simple Variant)}
\label{prompt:self_criteria}
\small
When no external rubric is provided (\texttt{use\_rubric=False}), the reviewer generates criteria using the following template injected into \texttt{\{rubric\}}:

\begin{quote}
\textbf{Self-Generated Evaluation Criteria:}\\
No external rubric is provided. You must generate evaluation criteria by combining standard research quality checks with problem-specific requirements.

Generate exactly 10 criteria in two parts:

\textbf{PART A} --- Standard research quality (4 criteria):
\begin{enumerate}
    \item Does the solution include a concrete experimental validation plan with specific datasets, baselines, or benchmarks appropriate for this problem?
    \item Is the proposed method technically sound and implementable as described?
    \item Is the methodology complete and detailed enough to execute end-to-end?
    \item Does the solution address the key constraints and practical considerations stated in the problem?
\end{enumerate}

\textbf{PART B} --- Problem-specific requirements (6 criteria):\\
Read the problem statement carefully and extract 6 criteria that are UNIQUE to this specific problem. Each must reference a concrete element from the problem (a named method, technique, dataset, domain term, metric, or constraint). These criteria should NOT apply to any other research problem.

Evaluate the proposed solution against ALL 10 criteria.
\end{quote}
\end{promptbox}

\begin{promptbox}{Rubric-Aware Oracle Condition}
\label{prompt:rubric_aware}
\small
When ground-truth rubrics are provided (\texttt{use\_rubric=True}), the following is injected into \texttt{\{rubric\}}:
\begin{quote}
\textbf{Evaluation Criteria (Rubric):}
\begin{itemize}
    \item \textit{\{ground-truth rubric item 1\}}
    \item \textit{\{ground-truth rubric item 2\}}
    \item ...
\end{itemize}
For EACH rubric item above, assess whether the proposed solution adequately addresses it.
\end{quote}
\end{promptbox}

\begin{promptbox}{Few-Shot Example Formatting}
\label{prompt:few_shot}
\small
For few-shot conditions, $k$(=5, 10, 15) examples are sampled from the ResearchPlanGen training split (fixed seed=42) and formatted as follows.

\textbf{For the Solver} (injected into the solver's \texttt{\{few\_shot\}}):
\begin{quote}
Here are examples of research scenarios with their evaluation criteria and expert solutions:

\textbf{Example 1:}\\
Scenario: \textit{\{Goal\}}\\
Evaluation Criteria:\\
--- \textit{\{rubric item 1\}}\\
--- \textit{\{rubric item 2\}} ...\\
Expert Solution: \textit{\{Reference solution\}}

\textit{[Repeated for $k$ examples]}

Now, given the following new research scenario, provide your research plan:
\end{quote}

\textbf{For the Reviewer} (injected into the reviewer's \texttt{\{few\_shot\}}):
\begin{quote}
Here are examples of evaluation criteria for similar research problems:

\textbf{Problem 1:} \textit{\{Goal, first 200 chars\}}...\\
Criteria:\\
--- \textit{\{rubric item 1\}}\\
--- \textit{\{rubric item 2\}} ...

\textit{[Repeated for $k$ examples]}

Now generate evaluation criteria for the current problem following similar patterns:
\end{quote}
\end{promptbox}

\begin{promptbox}{RPG Solver Prompt}
\label{prompt:rpg_solver}
\small
\textbf{System Role:}
\begin{quote}
You are an expert AI research scientist. Your task is to invent a plausible technical approach that could solve a given scientific problem in machine learning. You must be creative and innovative, proposing novel solutions.
\end{quote}
\textbf{User Prompt:}
\begin{quote}
I will provide you a research scenario. You have to provide me a concise yet thoughtful research plan with all details needed to execute it.

\texttt{\{few\_shot\}} \textit{[empty or $k$ formatted examples]}

\textbf{Here is the research scenario.}\\
Scenario: \texttt{\{problem\}}\\
\texttt{\{rubric\}} \textit{[self-generated or oracle criteria]}

\textbf{Overall Solution Guidelines:}
\begin{itemize}
    \item The plan should address the goals of the scenario and account for all constraints.
    \item Do NOT just say WHAT you will do. Explain HOW and WHY.
    \item Do not claim to have done experiments or have results, just provide the plan.
\end{itemize}

\textbf{Output:} Put the final solution inside \texttt{<solution></solution>} tags. Maximum 750 words.
\end{quote}
\end{promptbox}


\end{document}